\documentclass[10pt,twocolumn,letterpaper]{article}

\usepackage{iccv}
\usepackage{times}
\usepackage{epsfig}
\usepackage{graphicx}
\usepackage{amsmath}
\usepackage{amssymb}

\usepackage{multirow}
\usepackage{diagbox}
\usepackage{amsfonts} 

\usepackage{url}            
\usepackage{booktabs}       
\usepackage{amsfonts}       
\usepackage{nicefrac}       
\usepackage{microtype}      

\usepackage{times}
\usepackage{epsfig}
\usepackage{graphicx}
\usepackage{amsmath}
\usepackage{amssymb}
\usepackage{caption}
\usepackage{multirow}
\usepackage{bm}
\usepackage{algorithm, algpseudocode}
\usepackage{color}
\usepackage{colortbl}
\usepackage{textcomp}
\usepackage[outercaption]{sidecap}
\usepackage{makecell}
\usepackage{stmaryrd}
\usepackage{wrapfig}
\usepackage{caption}
\usepackage{subfigure}
\usepackage[table]{xcolor}
\usepackage[figuresleft]{rotating}
\makeatletter\renewcommand\paragraph{\@startsection{paragraph}{4}{\z@}{.5em \@plus1ex \@minus.2ex}{-.5em}{\normalfont\normalsize\bfseries}}\makeatother

\newcommand{\dec}[1]{\ensuremath{_{\text{\textcolor{blue}{($\downarrow$)}}}}}
\newcommand{\inc}[1]{\ensuremath{_{\text{\textcolor{magenta}{($\uparrow$)}}}}}
\newcommand{\grayback}[1]{\ensuremath{_{\text{\textcolor{gray! 20}{($\uparrow$)}}}}}
\newcommand{\wpic}[1]{\ensuremath{_{\text{\textcolor{white}{($\uparrow$)}}}}}

\usepackage[pagebackref=true,breaklinks=true,letterpaper=true,colorlinks,bookmarks=false]{hyperref}

\iccvfinalcopy 

\def\iccvPaperID{****} 

\ificcvfinal\fi

\begin{document}

\title{Set-level Guidance Attack: Boosting Adversarial Transferability of Vision-Language Pre-training Models}

\newcommand*\samethanks[1][\value{footnote}]{\footnotemark[#1]}
\author{Dong Lu${}^{1}$\thanks{$~$Equal contribution. ${}^\dagger$ Corresponding author.}, Zhiqiang Wang${}^1$\samethanks, Teng Wang${}^{1,2}$, Weili Guan${}^3$, Hongchang Gao${}^4$, Feng Zheng${}^{1,5\dagger}$\\
{\small ${}^1$Southern University of Science and Technology ${}^2$The University of Hong Kong}\\
{\small ${}^3$Monash University ${}^4$Temple University ${}^5$Peng Cheng Laboratory}\\
{\tt\small sammylu\_@outlook.com wangzq\_2021@outlook.com tengwang@connect.hku.hk}\\
{\tt\small honeyguan@gmail.com  hongchang.gao@temple.edu f.zheng@ieee.org}\\
}


\maketitle
\ificcvfinal\thispagestyle{empty}\fi

\begin{abstract}
Vision-language pre-training (VLP) models have shown vulnerability to adversarial examples in multimodal tasks. 
Furthermore, malicious adversaries can be deliberately transferred to attack other black-box models.
However, existing work has mainly focused on investigating white-box attacks.
In this paper, we present the first study to investigate the adversarial transferability of recent VLP models.
We observe that existing methods exhibit much lower transferability, compared to the strong attack performance in white-box settings.
The transferability degradation is partly caused by the under-utilization of cross-modal interactions.
Particularly, unlike unimodal learning, VLP models rely heavily on cross-modal interactions and the multimodal alignments are many-to-many, \textit{e.g.}, an image can be described in various natural languages.
To this end, we propose a highly transferable Set-level Guidance Attack (SGA) that thoroughly leverages modality interactions and incorporates alignment-preserving augmentation with cross-modal guidance.
Experimental results demonstrate that SGA could generate adversarial examples that can strongly transfer across different VLP models on multiple downstream vision-language tasks. 
On image-text retrieval, SGA significantly enhances the attack success rate for transfer attacks from ALBEF to TCL by a large margin (at least 9.78\% and up to 30.21\%), compared to the state-of-the-art.
\end{abstract}

\section{Introduction}
\label{sec:introduction}

Recent work has shown that vision-language pre-training (VLP) models are still vulnerable to adversarial examples \cite{Zhang2022Co-attack}, even though they have achieved remarkable performance on a wide range of multimodal tasks \cite{Shi2021DenseCV, Khan2021ExploitingBF, Lei2021UnderstandingCV}.
Existing work mainly focuses on white-box attacks, where information about the victim model is accessible.
However, the transferability of adversarial examples across VLP models has not been investigated, which is a more practical setting. 
It is still unknown whether the adversarial data generated on the source model can successfully attack another model, which poses a serious security risk to the deployment of VLP models in real-world applications. 

\begin{figure}
\begin{center}
   \includegraphics[width=0.97\linewidth]{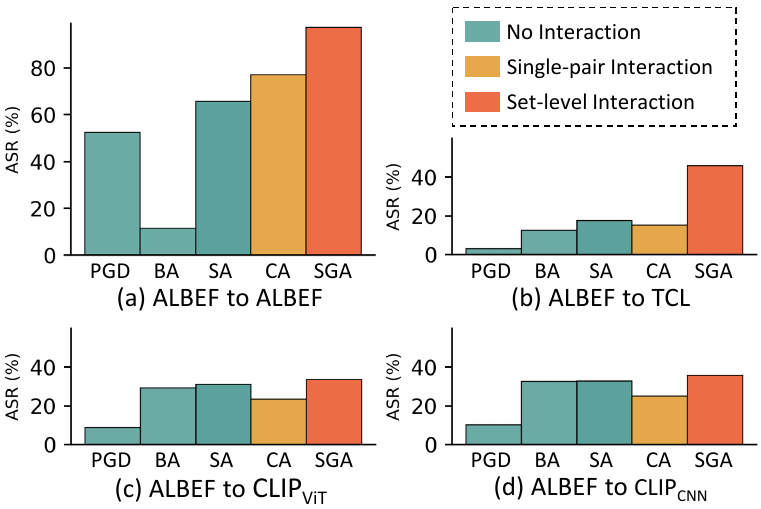}
\end{center}
\vspace{-11pt}
   \caption{
\textbf{Comparison of attack success rates (ASR) using five different attacks on image-text retrieval.}
Adversarial examples are crafted on the source model (ALBEF) to attack the target white-box model or black-box models.
The first three columns refer to the image-only \textbf{PGD} attack \cite{Madry2018PGD}, text-only BERT-Attack \cite{Li2020BERTATTACK} (\textbf{BA}), and the combined separate unimodal attack (\textbf{SA}), which all belong to the methods without cross-modal interactions.
The fourth column is the state-of-the-art multimodal Co-Attack \cite{Zhang2022Co-attack} (\textbf{CA}) that employs single-pair cross-modal interactions.
The last column is the proposed Set-level Guidance Attack (\textbf{SGA}), which leverages multiple set-level cross-modal interactions, successfully attacking the white-box model and transferring to attack all black-box models with the highest ASR. 
More discussions are in Section \ref{sec:analysis}.
}
\vspace{-8pt}
\label{fig:f1_cover}
\end{figure}

This paper makes the first step to investigate the transferability of adversarial samples within VLP models. 
Without loss of generality, most of our experiments are based on image-text retrieval tasks. 
We first empirically evaluate this attack performance with respect to different modalities on multimodal tasks across multiple datasets. 
Our results show that the adversarial transferability of attacking both modalities (image \& text) consistently beats attacking unimodal data (image or text). 
Unfortunately, even though two modalities are allowed to be perturbed simultaneously, the attack success rates of existing methods \cite{Madry2018PGD, Li2020BERTATTACK, Zhang2022Co-attack} still significantly drops when transferring from the white-box to black-box settings, as shown in Figure \ref{fig:f1_cover}.

Different from recent studies focusing on separate attacks on unimodal data \cite{Madry2018PGD, Li2020BERTATTACK}, multimodal pairs exhibit intrinsic alignment and complementarity to each other.
The modeling of inter-modal correspondence turns out to be a critical problem for transferability. 
Considering that the alignments between image and text are many-to-many, for example, an image could be described to be with various human perspectives and language styles, a reasonable perturbation direction may be determined with diverse guidance from multiple labels in the other modality. 
However, recent adversarial attack methods for VLP models \cite{Zhang2022Co-attack} usually employ a single image-text pair to generate adversarial samples. 
Although they exhibit strong performance in white-box settings, the poor diversity of guidance makes adversarial samples highly correlated with the alignment pattern of the white-box model, and therefore impedes generalization to black-box settings.

\begin{figure}
\begin{center}
   \includegraphics[width=0.99\linewidth]{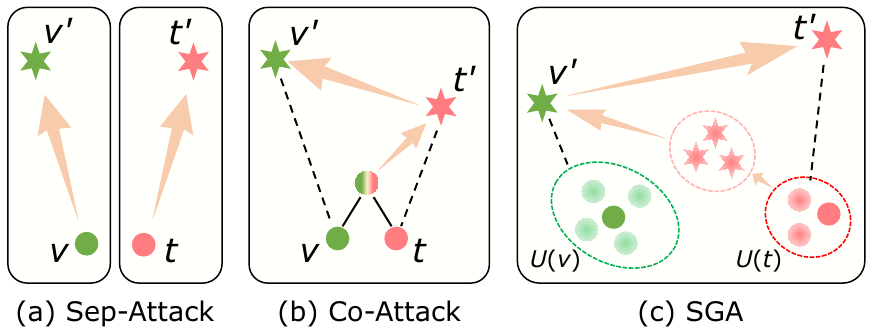}
\end{center}
\vspace{-12pt}
   \caption{
\textbf{Comparison of cross-modal interactions.}
To generate adversarial examples, existing methods either: \textbf{(a)} separately perturb unimodal data without any cross-modal interactions (\textbf{Sep-Attack}) or, \textbf{(b)} perturb multiple modalities but with single image-text pairs to model cross-modal interactions (\textbf{Co-Attack}).
However, our method is capable of learning cross-modal interactions among multiple alignments through \textbf{(c)} set-level guidance (\textbf{SGA}). 
Specifically, $v$ denotes the input image, and $t$ is the paired caption.
$v'$ and $t'$ represent the corresponding adversarial examples, respectively.
$U(v)$ and $U(t)$ represent the 
scale-invariant image set and most matching caption set.
Arrows indicate the guidance for generating adversarial examples. 
}
\vspace{-10pt}
\label{fig:f2_SGA}
\end{figure}

To address the weak transferability problem, we propose Set-level Guidance Attack (SGA), which leverages diverse cross-modal interactions among multiple image-text pairs (Figure \ref{fig:f2_SGA}). 
Specifically, we introduce alignment-preserving augmentation which enriches image-text pairs while keeping their alignments intact. 
The image augmentation is based on the scale-invariant property of deep learning models \cite{Lin2019NesterovScaleInva}, thus we can construct multi-scale images to increase the diversity. For text augmentation, we select the most matching caption pairs from the dataset.
More importantly, SGA generates adversarial examples on multimodal augmented input data with carefully designed cross-modal guidance. 
In detail, SGA iteratively pushes supplemental information away between two modalities with another modality as supervision to disrupt the interactions for better harmonious perturbations. 
Note that resultant adversarial samples could perceive the gradients originated from multiple guidance.

We conduct experiments on two well-established multimodal datasets, Flickr30K \cite{Plummer2015Flickr30k} and MSCOCO \cite{Lin2014COCO}, to evaluate the performance of our proposed SGA across various Vision-and-Language (V+L) downstream tasks.
The experimental results demonstrate the high effectiveness of SGA in generating adversarial examples that can be strongly transferred across VLP models, surpassing the current state-of-the-art attack methods in multimodal learning. 
In particular, SGA achieves notable improvements in image-text retrieval under black-box settings and also exhibits superior performance in white-box attack settings.
Moreover, SGA also outperforms the state-of-the-art methods in image captioning and yields higher fooling rates on visual grounding. 

We summarize our contributions as follows. 
\textbf{1)} We make the first attempt to explore the transferability of adversarial examples on popular VLP models with a systematical evaluation; 
\textbf{2)} We provide SGA, a novel transferable multimodal attack that enhances adversarial transferability through the effective use of set-level alignment-preserving augmentations and well-designed cross-modal guidance;
\textbf{3)} Extensive experiments show that SGA consistently boosts adversarial transferability across different VLP models than the state-of-the-art methods.
\section{Related Work}
\subsection{Vision-Language Pre-training Models}
Vision-Language Pre-training (VLP) aims to improve the performance of downstream multimodal tasks by pre-training large-scale image-to-text pairs \cite{Li2022BLIP}.
Most works are developed upon the pre-trained object detectors with region features to learn the vision-language representations \cite{Chen2020UNITER,Li2020Oscar,Zhang2021VinVL,wang2022vlmixer}. 
Recently, with the increasing popularity of Vision Transformer (ViT) \cite{Dosovitskiy2021ViT, Touvron2021TrainingDI_forBLIP_arhc,  Yuan2021TokenstoTokenVT}, some other works propose to use ViT as an image encoder and transform the input into patches in an end-to-end manner \cite{Li2021ALBEF,Yang2022TCL,Li2022BLIP,Dou2021METER,wang2023accelerating}.

According to the VLP architectures, VLP models can be classified into two typical types: fused VLP models and aligned VLP models \cite{Zhang2022Co-attack}.  
Specifically, fused VLP models (\textit{e.g.}, ALBEF \cite{Li2021ALBEF}, TCL \cite{Yang2022TCL}) first utilize separate unimodal encoders to process token embeddings and visual features, and further use a multimodal encoder to process image and text embeddings to output fused multimodal embeddings.
Alternatively, aligned VLP models (\textit{e.g.}, CLIP \cite{Radford2021CLIP}) have only unimodal encoders with independent image and text modality embeddings.
In this paper, we focus on popular architectures with fused and aligned VLP models.

\subsection{Image-Text Retrieval Task}
Image-Text Retrieval (ITR) aims to retrieve the relevant top-ranked instances from a gallery database with one modality, given an input query from another modality \cite{Wang2019CAMP, Chen2020IMRAM, Zhang2020ContextAwareAN, Cheng2022ViSTAVA}.
This task can be divided into two subtasks, image-to-text retrieval (TR) and text-to-image retrieval (IR).

For ALBEF \cite{Li2021ALBEF} and TCL \cite{Yang2022TCL}, the semantic similarity score in the unimodal embedding space will be calculated for all image-text pairs to select top-$k$ candidates. Then the multimodal encoder takes the top-$k$ candidates and computes the image-text matching score for ranking.
For CLIP \cite{Radford2021CLIP}, without the multimodal encoder, the final rank list can be obtained based on the similarity in the embedding space between image and text modalities.

\subsection{Adversarial Transferability}
Existing adversarial attacks can be categorized into two settings: white-box attacks and black-box attacks. 
In a white-box setting, the target model is fully accessible, but not in a black-box setting.
In computer vision, many methods employ gradient information for adversarial attacks in white-box settings, such as FGSM \cite{Goodfellow2015FGSM}, PGD \cite{Madry2018PGD}, C\&W \cite{Carlini2017CW}, and MIM \cite{Dong2018BoostingAA}.
In contrast, in the field of natural language processing (NLP), current attack methods mainly modify or replace some tokens of the input text \cite{Li2020BERTATTACK, Ren2019NLP_Adv_Saliency, Gao2018BlackBoxGO, Jin2020IsBR}.
In the multimodal vision-language domain, Zhang \textit{et al.} \cite{Zhang2022Co-attack} proposed a white-box multimodal attack method with respect to popular VLP models on downstream tasks. 

However, white-box attacks are unrealistic due to the inaccessibility of model information in practical applications.
In addition, there is no related work that systematically analyzes the adversarial transferability of multimodal attack methods on VLP models.
Therefore, in this work, we mainly focus on generating highly transferable adversaries across different VLP models.

\section{Analysis of Adversarial Transferability}
\label{sec:analysis}

\begin{figure}[t]
\begin{center}
   \includegraphics[width=0.95\linewidth]{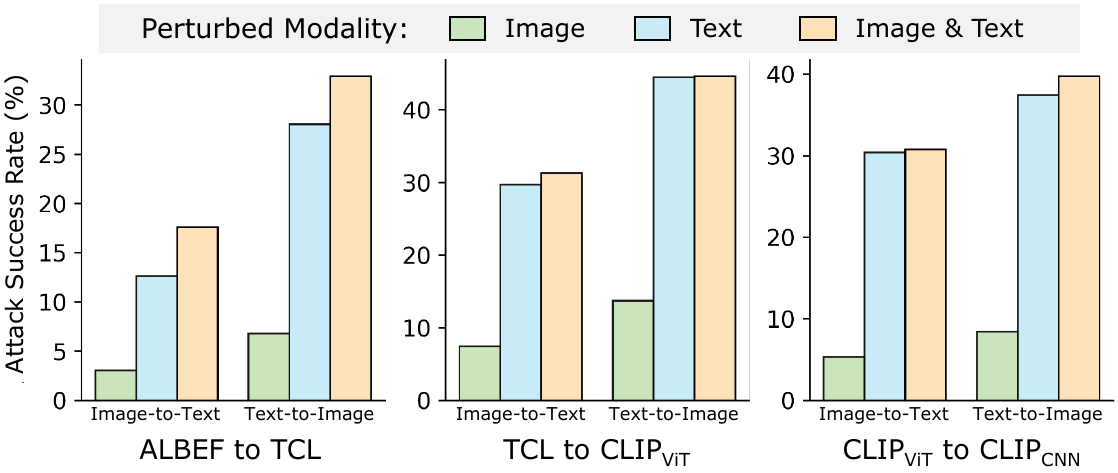}
\end{center}
\vspace{-11pt}
   \caption{\textbf{Attack success rates ($\%$) of different perturbed modalities on image-text retrieval.} Adversarial examples with respect to different modalities from the source model to attack the target model using Sep-Attack. Different colors represent different perturbed modalities.}
\label{fig:f3_ana_perturb_input}
\end{figure}

\begin{figure}
\begin{center}
   \includegraphics[width=0.95\linewidth]{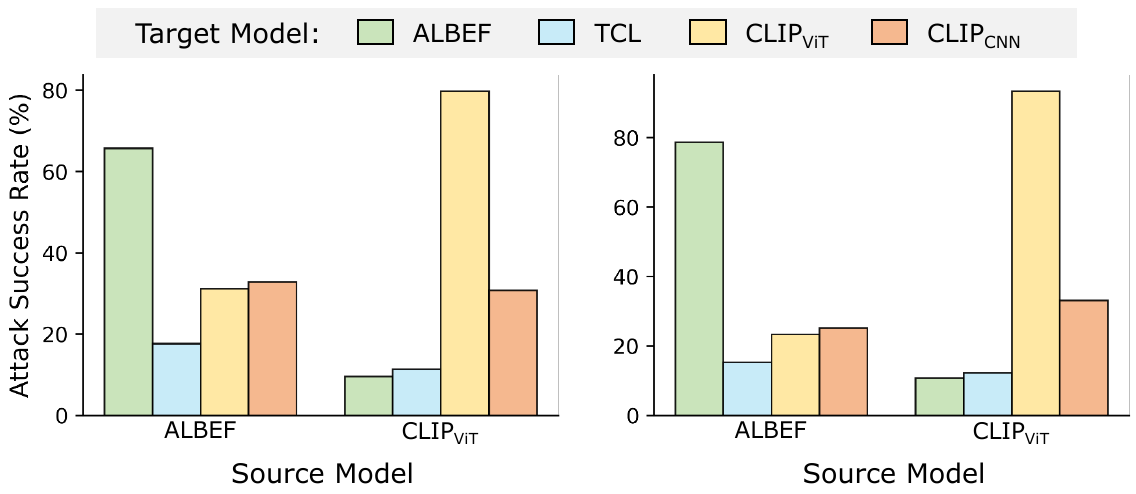}
\end{center}
\vspace{-11pt}
   \caption{\textbf{Attack success rates ($\%$) on white-box models and black-box models.} Both images and text are perturbed by Sep-Attack (\textbf{Left}) and Co-Attack (\textbf{Right}) on image-text retrieval. Colors indicate different target models.}
   \vspace{-10pt}
\label{fig:f4_ana_transfer}
\end{figure}

In this section, we conduct an empirical study on VLP models to evaluate adversarial transferability using existing methods.
A common approach to attack multimodal tasks is combining unimodal adversarial attacks \cite{Madry2018PGD, Dong2018BoostingAA, Xie2018ImprovingTransDiverse, Li2020BERTATTACK} of each modalities together. 
For instance, the separate unimodal attack (Sep-Attack) includes PGD \cite{Madry2018PGD} and BERT-Attack \cite{Li2020BERTATTACK} for attacking image modality and text modality, respectively.
Other recent multimodal adversarial attacks, such as Co-Attack \cite{Zhang2022Co-attack}, consider cross-modal interactions by perturbing image modalities and text modalities collectively.

We first present the observations regarding the adversarial transferability of VLP models.
Then we discuss the limitations of the existing methods.
By conducting this study, we aim to provide insights into the robustness of VLP models against adversarial attacks and the effectiveness of different attack strategies.

\subsection{Observations}
To investigate the adversarial transferability of perturbed inputs with respect to different modalities (\textit{i.e.}, image, text, and image \& text) and the effect of different VLP models on transferability, we conduct experiments and present the attack success rates of the adversarial examples generated by the source model to attack the target models in Figure \ref{fig:f3_ana_perturb_input} and Figure \ref{fig:f4_ana_transfer}.
The observations are summarized below:
\begin{itemize}
  \item The adversarial transferability of attacking both modalities (image \& text) is consistently more effective than attacking any unimodal data alone (image or text).
  As shown in Figure \ref{fig:f3_ana_perturb_input}, transferring both adversarial image and text from ALBEF to TCL leads to a much higher attack success rate than transferring adversarial examples of any single modality. Notably, ALBEF and TCL are both fused VLP models.
  Similar observations also exist in the following settings: (1) The source and target models are different types of VLP models but have the same basic architectures (\textit{e.g.}, TCL and CLIP$_{\rm ViT}$). (2) The source and target models are the same types of VLP models, but with different basic architectures (\textit{e.g.}, CLIP$_{\rm ViT}$ and CLIP$_{\rm CNN}$).
  \item Adversarial multimodal data (\textit{i.e.}, adversarial image \& text), which have strong attack performance on the source model, can hardly maintain the same capability when transferring to target models. 
  For example, as illustrated in Figure \ref{fig:f4_ana_transfer}, even though ALBEF and TCL have the same model architecture, the attack success rate sees a significant drop when transferring adversarial examples generated on ALBEF to TCL.
  The phenomenon exists in both Sep-Attack and Co-Attack.
\end{itemize}

In summary, the attack methods can have stronger transferability in black-box settings if all the modalities are attacked simultaneously.
However, even though two modalities are allowed to be perturbed, existing methods still exhibit much lower transferability. 
This suggests that attacks with higher transferability should be specifically designed, instead of directly utilizing the existing white-box attack methods.

\subsection{Discussions}\label{sec:sec3_dis}
We posit that the degradation in transferability of adversarial examples is mainly due to the limitations of the existing attack methods:
\begin{itemize}
  \item One major limitation of Sep-Attack is that it does not take into account the interactions between different modalities. 
  As a combined independent attack method for each modality, it fails to model the inter-modal correspondence that is crucial for successful attacks in multimodal learning.
  This is particularly evident in multimodal tasks such as image-text retrieval, where the ground truth is not discrete labels (\textit{e.g.}, image classification) but another modality data that corresponds to the input modality.
  The complete lack of cross-modal interactions in Sep-Attack severely limits the generalization of adversarial examples and reduces their transferability among different VLP models. 
  \item While Co-Attack is designed to leverage the collaboration between modalities to generate adversarial examples, it still suffers from a key drawback that hinders its transferability to other VLP models.
  Unlike unimodal learning, multimodal learning involves multiple complementary modalities with many-to-many cross-modal alignments, which pose unique challenges to achieving sufficient adversarial transferability.
  However, Co-Attack only uses single image-text pairs to generate adversarial data, limiting the diversity of guidance from multiple labels in other modalities.
  This lack of diversity in cross-modal guidance makes adversarial samples highly correlated with the alignment pattern of the white-box model.
  Therefore, the generality of adversarial examples is restricted, and their effectiveness in transferring to other models drops.
\end{itemize}

In conclusion, the analysis motivates our investigation into the adversarial transferability of VLP models.
Moreover, it highlights the pressing need to explore transferable multimodal attacks for generating adversarial examples that can be effectively transferred across different VLP models.
\section{Methodology}
\label{sec:method}
In this section, we propose a transferable multimodal adversarial attack, termed Set-level Guidance Attack (SGA).
SGA is designed to enhance adversarial transferability across VLP models by leveraging multimodal interactions and incorporating diverse alignment-preserving inter-modal information with carefully designed cross-modal guidance.
We provide our motivation first, followed by relevant notations, and finally present SGA in detail.

\subsection{Motivation}
To improve the transferability of multimodal attacks, we first conduct an investigation into the shortcomings of existing methods in black-box settings. 
Through a systematic analysis of failure cases, we observe that around half of these cases arise due to the presence of multiple matched captions of the image, as shown in Table \ref{tab:t1_black_box_failures}. 
More specifically, our findings indicate that while the generated adversarial image may be significantly distant from the single supervised caption in the source model, it is prone to approaching other matched captions in the target model, where the alignments can be modeled and ultimately lead to attack failure.

Therefore, to maintain the attack ability of adversarial images when transferring to other models, it is crucial to consider multiple paired captions and push the adversarial image away from all the paired captions, thus preserving the attack ability when transferred to other black-box models.
Crafting adversarial text for high transferability follows a similar approach, which can also benefit from more paired images.
More discussions can be found in Appendix \ref{sec:supp_a}.

\begin{table}[t]
\begin{center}
\footnotesize
\renewcommand\arraystretch{1}
\setlength{\tabcolsep}{2.2pt}
		\scalebox{0.97}[0.97]{
		\begin{tabular}{  l|c|c|ccc}
        \toprule[0.3mm]
			&  &{$p$ (\textit{Event A})} $\uparrow$ & \multicolumn{3}{c}{$p$ (\textit{Event B} $|$\textit{ Event A}) $\downarrow$} \\
			\cmidrule{3-6}
			\multirow{-2}{*}{\textbf{Attack}} & \multirow{-2}{*}{\makecell[c]{Cross-modal \\ Interaction}}& {ALBEF} & {TCL} & {CLIP$_{\rm ViT}$} & {CLIP$_{\rm CNN}$}\\
			\midrule
            Sep-Attack  &  No  & 74.60\%    & 46.78\%   & 41.69\%   & 41.82\%   \\
            Co-Attack   &  Single-pair  & 80.60\%    & 50.50\%   & 40.82\%   & 42.93\%   \\
            SGA         &  Set-level     & \textbf{97.20\%}    & \textbf{28.91\%}  & \textbf{34.67\%}   & \textbf{38.58\%}\\
			\bottomrule[0.3mm]
	\end{tabular}}
\end{center}
\vspace{-8pt}
\caption{ 
\textbf{Adversarial images with insufficient cross-modal interactions may get closer to other paired captions when transferring.} 
\textit{\textbf{Event A}}: success cases in white-box settings.
\textit{\textbf{Event B}}: failure cases in black-box settings influenced by other paired captions.
Adversarial data are generated on Flickr30K from ALBEF to attack other target models.
}
\vspace{-10pt}
\label{tab:t1_black_box_failures}
\end{table}

\subsection{Notations}
Let $(v,t)$ denote an image-text pair sampled from a multimodal dataset $D$. 
For VLP models, we denote $f_I$ as the image encoder and $f_T$ as the text encoder. 
The multimodal fusion module in fused VLP models is denoted by $f_M$.
Specifically, $f_I(v)$ represents the image representation $e_v$ encoded by $f_I$ taking an image $v$ as input, $f_T(t)$ denotes the text representation $e_t$ encoded by $f_T$ taking a text $t$ as input, and $f_M(e_v,e_t)$ denotes the multimodal representation encoded by $f_M$ taking image and text representations as inputs.

We use $B[v,\epsilon_v]$ and $B[t,\epsilon_t]$ to represent the legal searching spaces for optimizing adversarial image and text, respectively.
Specifically, $\epsilon_v$ denotes the maximal perturbation bound for the image, and $\epsilon_t$ denotes the maximal number of changeable words in the caption.

\subsection{Transferable Set-Level Guidance Attack}
\paragraph{Alignment-preserving Augmentation.}
The analysis presented in Section \ref{sec:analysis} highlights the key limitation of existing methods: the inter-modal information used to generate adversarial examples lacks diversity.
The limitation will make the generated adversarial examples fail to generalize to other black-box models with strong attack performance, resulting in limited adversarial transferability.

To inject more diversity in the generation of generalizable adversarial examples, we propose using set-level alignment-preserving augmentation to expand multimodal input spaces while maintaining cross-modal alignments intact.
Unlike previous methods that only consider a single image-text paired example $(v,t)$ to generate adversarial data, we enlarge the input to a set level of images and captions.
Specifically, we select the most matching caption pairs from the dataset of each image $v$ to form an augmented caption set $\boldsymbol{t}=\{t_1, t_2,...,t_M\}$, and resize each image $v$ into different scales $S=\{s_1,s_2,...,s_N\}$ and then add Gaussian noise to obtain a multi-scale image set $\boldsymbol{v} = \{v_1,v_2,...,v_N\}$ based on the scale-invariant property. 
The enlarged input set $(\boldsymbol{v},\boldsymbol{t})$ is then used to generate the adversarial data $(v',t')$.

\paragraph{Cross-modal Guidance.}
Cross-modal interactions play a crucial role in multimodal tasks. 
For example, in image-text retrieval, the paired information from another modality provides unique annotation supervision for each sample. Similarly, in adversarial attacks, supervisory information is essential in guiding the search for adversarial examples.

To fully utilize the enlarged alignment-preserving multimodal input set $(\boldsymbol{v},\boldsymbol{t})$ and further improve the transferability of the generated adversarial data, we propose cross-modal guidance to utilize interactions from different modalities. Specifically, we use the paired information from another modality as the supervision to guide the direction of optimizing the adversarial data.
This guidance iteratively pushed away the multimodal information and disrupt the cross-modal interaction for better harmonious perturbations. 
Notably, the resultant adversarial examples can perceive the gradients originated from multiple guidance.

First, we generate corresponding adversarial captions for all captions in the text set $\boldsymbol{t}$, forming an adversarial caption set $\boldsymbol{t'}=\{t_{1}', t_{2}'...,t_{M}'\}$. 
The process can be formulated as,
\begin{equation}\label{eq:CMAG_t_group}
  t_{i}' = \mathop{\arg\max}\limits_{t_{i}'\in {B}[t_i,\epsilon_t]}-\frac{f_T(t_{i}')\cdot f_I(v)}{\Vert f_T(t_{i}') \Vert \Vert f_I(v) \Vert}.
\end{equation}
The adversarial caption $t_{i}'$ is constrained  to be dissimilar to the original image $v$ in the embedding space. 
Next, the adversarial image $v'$ is generated by solving
\begin{equation}\label{eq:CMAG_x_adv}
\small
  v'=\mathop{\arg\max}\limits_{v'\in {B}[v,\epsilon_v]} -\sum_{i=1}^{M}\frac{f_T(t_{i}')}{\Vert f_T(t_{i}')\Vert}\sum_{s_i\in S}\frac{f_I(g(v', s_i))}{\Vert f_I(g(v', s_i))\Vert},
\end{equation}
where $g(v',s_i)$ denotes the resizing function that takes the image $v'$ and the scale coefficient $s_i$ as inputs. 
All the scaled images derived from $v'$ are encouraged to be far away from all the adversarial captions $t_{i}'$ in the embedding space.
Finally, the adversarial caption $t'$ is generated as follows,
\begin{equation}\label{eq:CMAG_t_adv}
\small
  t' = \mathop{\arg\max}\limits_{t'\in {B}[t,\epsilon_t]}-\frac{f_T(t')\cdot f_I(v')}{\Vert f_T(t') \Vert \Vert f_I(v') \Vert},
\end{equation}
in which $t'$ is encouraged to be far away from the adversarial image $v'$ in the embedding space. 
The detailed algorithm can be found in Appendix \ref{sec:supp_c}.

\begin{table*}[t]
\begin{center}
\small
\renewcommand\arraystretch{1.1}
\setlength{\tabcolsep}{3.5pt}
		\scalebox{0.97}[0.97]{
		\begin{tabular}{  l|cc|cc|cc|cc}
        \toprule[0.3mm]
			&  \multicolumn{2}{c}{\textbf{ALBEF$^\ast$}} & \multicolumn{2}{c}{\textbf{TCL}} & \multicolumn{2}{c}{\textbf{CLIP$_{\rm ViT}$}} & \multicolumn{2}{c}{\textbf{CLIP$_{\rm CNN}$}}  \\
			\cmidrule{2-9}
			\multirow{-2}{*}{\textbf{Attack}} & {TR R@1$^\ast$} & {IR R@1$^\ast$} & {TR R@1} & {IR R@1} & {TR R@1} & {IR R@1} & {TR R@1} & {IR R@1}  \\
			\midrule
           Sep-Attack & 65.69\wpic{65.69} & 73.95\wpic{73.95} & 17.60\wpic{17.60} & 32.95\wpic{32.95} & 31.17\wpic{31.17} &  \textbf{45.23}\wpic{45.23} & 32.82\wpic{32.82} & 45.49\wpic{45.49} \\
            \     Sep-Attack + MI & 58.81\dec{6.88} & 65.25\dec{8.7} & 16.02\dec{1.58} & 28.19\dec{4.76} & 23.07\dec{8.1} & 36.98\dec{8.25} & 26.56\dec{6.26} & 39.31\dec{6.18} \\
            \     Sep-Attack + DIM & 56.41\dec{9.28} & 64.24\dec{9.71} & 16.75\dec{0.85} & 29.55\dec{3.4} & 24.17\dec{7.0} & 37.60\dec{7.63} & 25.54\dec{7.28} & 38.77\dec{6.72}  \\
            \     Sep-Attack + PNA\_PO & 40.56\dec{25.13} & 53.95\dec{20.0} & 18.44\inc{0.84} & 30.98\dec{1.97} & 22.33\dec{8.84} & 37.02\dec{8.21} & 26.95\dec{5.87} & 38.63\dec{6.86} \\
            \midrule
            Co-Attack & 77.16\wpic{77.16} & 83.86\wpic{83.86} & 15.21\wpic{15.21} & 29.49\wpic{29.49} & 23.60\wpic{23.60} & 36.48\wpic{36.48} & 25.12\wpic{25.12} & 38.89\wpic{38.89} \\
            \     Co-Attack + MI & 64.86\dec{12.3} & 75.26\dec{8.6} & 25.40\inc{10.19} & 38.69\inc{9.2} & 24.91\inc{1.31} & 37.11\inc{0.63} & 26.31\inc{1.19} & 38.97\inc{0.08} \\
            \     Co-Attack + DIM & 47.03\dec{30.13} & 62.28\dec{21.58} & 22.23\inc{7.02} & 35.45\inc{5.96} & 25.64\inc{2.04} & 38.50\inc{2.02} & 26.95\inc{1.83} & 40.58\inc{1.69} \\
            \midrule
            \rowcolor{gray! 20} SGA & \textbf{97.24}\grayback{2} & \textbf{97.28}\grayback{2} & \textbf{45.42}\grayback{2} & \textbf{55.25}\grayback{2} & \textbf{33.38}\grayback{2} & 44.16\grayback{2} & \textbf{34.93}\grayback{2} & \textbf{46.57}\grayback{2} \\
			\bottomrule[0.3mm]
	\end{tabular}}
\end{center}
\vspace{-8pt}
\caption{\textbf{Attack success rates ($\%$) of R@1 of integrating transfer-based image attacks on image-text retrieval.} The adversarial data are generated on Flickr30K using the source model ALBEF to attack other target models. $^\ast$ indicates white-box attacks. A higher ASR indicates better adversarial transferability.}
\vspace{-12pt}
\label{tab:t2_exp_unimodal}
\end{table*}

\section{Experiments}
\label{sec:experiment}

In this section, we present experimental evidence for the advantages of our proposed SGA. 
We conduct experiments on a diverse set of datasets and popular VLP models.
First, we describe the experimental settings in Section \ref{sec: exp_setting}. 
Then, in Section \ref{sec: exp_ana}, we validate the immediate integration of transfer-based unimodal attacks into multimodal learning. 
Next, we provide the main evaluation results compared to the state-of-the-art method in Section \ref{sec: exp_res}. 
In Section \ref{sec:cross_task}, we analyze the cross-task transferability between different V+L tasks. 
Finally, we present ablation studies in Section \ref{sec: exp_ablation}.

\subsection{Experimental Settings}\label{sec: exp_setting}
\paragraph{Datasets.} We consider two widely used multimodal datasets, Flickr30K \cite{Plummer2015Flickr30k} and MSCOCO \cite{Lin2014COCO}.
Flickr30K consists of 31,783 images, each with five corresponding captions.
Similarly, MSCOCO comprises 123,287 images, and each image is annotated with around five captions.
We adopt the Karpathy split \cite{Karpathy2017KarpathySplit} for experimental evaluation. 

\paragraph{VLP Models.} We evaluate two popular VLP models, the fused VLP and aligned VLP models. 
For the fused VLP, we consider ALBEF \cite{Li2021ALBEF} and TCL \cite{Yang2022TCL}.
ALBEF contains a 12-layer visual transformer ViT-B/16 \cite{Dosovitskiy2021ViT} and two 6-layer transformers for the image encoder and both the text encoder and the multimodal encoder, respectively. 
TCL uses the same model architecture as ALBEF but with different pre-train objectives.
For the aligned VLP model, we choose to evaluate CLIP \cite{Radford2021CLIP}.
CLIP has two different image encoder choices, namely, $\rm CLIP_{ViT}$ and $\rm CLIP_{CNN}$, that use ViT-B/16 and ResNet-101 \cite{He2016ResNet} as the base architectures for the image encoder, respectively.

\paragraph{Adversarial Attack Settings.} 
To craft adversarial images, we employ PGD \cite{Madry2018PGD} with perturbation bound $\epsilon_{v}=2/255$, step size $\alpha=0.5/255$, and iteration steps $T=10$.
For attacking text modality, we adopt BERT-Attack \cite{Li2020BERTATTACK} with perturbation bound $\epsilon_{t}=1$ and length of word list $W=10$.
Furthermore, we enlarge the image set by resizing the original image into five scales, $\{0.50, 0.75, 1.00, 1.25, 1.50\}$, using bicubic interpolation.
Similarly, the caption set is enlarged by augmenting the most matching caption pairs for each image in the dataset, with the size of approximately five.

\paragraph{Metrics.}
We employ Attack Success Rate (ASR) as the metric for evaluating the adversarial robustness and transferability in both white-box and black-box settings. 
Specifically, ASR evaluates the percentage of attacks that only produce successful adversarial examples. 
A higher ASR indicates better adversarial transferability.

\begin{table*}[t]
\begin{center}
\small
\renewcommand\arraystretch{1}
\setlength{\tabcolsep}{2.5pt}
		\scalebox{0.95}[0.95]{
		\begin{tabular}{ @{\extracolsep{\fill}} l|l|cc|cc|cc|cc} 
        \toprule[0.3mm]
			& &  \multicolumn{2}{c}{\textbf{ALBEF}} & \multicolumn{2}{c}{\textbf{TCL}} & \multicolumn{2}{c}{\textbf{CLIP$_{\rm ViT}$}} & \multicolumn{2}{c}{\textbf{CLIP$_{\rm CNN}$}}  \\
			\cmidrule{3-10} 
			\multirow{-2}{*}{\textbf{Source}} &\multirow{-2}{*}{\textbf{Attack}} & {TR R@1} & {IR R@1} & {TR R@1} & {IR R@1} & {TR R@1} & {IR R@1} & {TR R@1} & {IR R@1}  \\
			\midrule
			\multirow{5}{*}{\rotatebox[origin=c]{0}{\textbf{ALBEF}}} 
            & PGD & 52.45$^\ast$ & 58.65$^\ast$ & 3.06 & 6.79 & 8.96 & 13.21 & 10.34 & 14.65 \\
            & BERT-Attack & 11.57$^\ast$ & 27.46$^\ast$ & 12.64 & 28.07 & 29.33 & 43.17 & 32.69 & 46.11 \\
            & Sep-Attack & 65.69$^\ast$ & 73.95$^\ast$ & 17.60 & 32.95 & 31.17 & \textbf{45.23} & 32.82 & 45.49\\
            & Co-Attack & 77.16$^\ast$ & 83.86$^\ast$ & 15.21 & 29.49 & 23.60 & 36.48 & 25.12 & 38.89  \\
			& \cellcolor{gray! 20}SGA  & \cellcolor{gray! 20}\textbf{97.24$\pm$0.22$^\ast$} & \cellcolor{gray! 20}\textbf{97.28$\pm$0.15$^\ast$} & \cellcolor{gray! 20}\textbf{45.42$\pm$0.60} & \cellcolor{gray! 20}\textbf{55.25$\pm$0.06} & \cellcolor{gray! 20}\textbf{33.38$\pm$0.35} & \cellcolor{gray! 20}44.16$\pm$0.25 & \cellcolor{gray! 20}\textbf{34.93$\pm$0.99} & \cellcolor{gray! 20}\textbf{46.57$\pm$0.13} \\
			\midrule 
			\multirow{5}{*}{\rotatebox[origin=c]{0}{\textbf{TCL}}} 
            & PGD  & 6.15 & 10.78 & 77.87$^\ast$ & 79.48$^\ast$ & 7.48 & 13.72 & 10.34 & 15.33 \\
            & BERT-Attack & 11.89 & 26.82 & 14.54$^\ast$ & 29.17$^\ast$ & 29.69 & 44.49 & 33.46 & 46.07 \\
            & Sep-Attack & 20.13 & 36.48 & 84.72$^\ast$ & 86.07$^\ast$ & 31.29 & 44.65 & 33.33 & 45.80\\
            & Co-Attack & 23.15 & 40.04 & 77.94$^\ast$ & 85.59$^\ast$ & 27.85 & 41.19 & 30.74 & 44.11 \\
			& \cellcolor{gray! 20}SGA  & \cellcolor{gray! 20}\textbf{48.91$\pm$0.74} & \cellcolor{gray! 20}\textbf{60.34$\pm$0.10} & \cellcolor{gray! 20}\textbf{98.37$\pm$0.08$^\ast$} & \cellcolor{gray! 20}\textbf{98.81$\pm$0.07$^\ast$} & \cellcolor{gray! 20}\textbf{33.87$\pm$0.18} & \cellcolor{gray! 20}\textbf{44.88$\pm$0.54} & \cellcolor{gray! 20}\textbf{37.74$\pm$0.27} & \cellcolor{gray! 20}\textbf{48.30$\pm$0.34} \\
			\midrule 
			\multirow{5}{*}{\rotatebox[origin=c]{0}{\textbf{CLIP$_{\rm ViT}$}}} 
            & PGD & 2.50 & 4.93 & 4.85 & 8.17 & 70.92$^\ast$ & 78.61$^\ast$ & 5.36 & 8.44  \\
            & BERT-Attack  & 9.59 & 22.64 & 11.80 & 25.07 & 28.34$^\ast$ & 39.08$^\ast$ & 30.40 & 37.43  \\
            & Sep-Attack & 9.59 & 23.25 & 11.38 & 25.60 & 79.75$^\ast$ & 86.79$^\ast$ & 30.78 & 39.76\\
            & Co-Attack & 10.57 & 24.33 & 11.94 & 26.69 & 93.25$^\ast$ & 95.86$^\ast$ & 32.52 & 41.82 \\
			& \cellcolor{gray! 20}SGA  & \cellcolor{gray! 20}\textbf{13.40$\pm$0.07} & \cellcolor{gray! 20}\textbf{27.22 $\pm$0.06} & \cellcolor{gray! 20}\textbf{16.23$\pm$0.45} & \cellcolor{gray! 20}\textbf{30.76$\pm$0.07} & \cellcolor{gray! 20}\textbf{99.08$\pm$0.08$^\ast$} & \cellcolor{gray! 20}\textbf{98.94$\pm$0.00$^\ast$} & \cellcolor{gray! 20}\textbf{38.76$\pm$0.27} & \cellcolor{gray! 20}\textbf{47.79$\pm$0.58} \\
			\midrule 
			\multirow{5}{*}{\rotatebox[origin=c]{0}{\textbf{CLIP$_{\rm CNN}$}}} 
            & PGD & 2.09 & 4.82 & 4.00 & 7.81 & 1.10 & 6.60 & 86.46$^\ast$ & 92.25$^\ast$\\
            & BERT-Attack & 8.86 & 23.27 & 12.33 & 25.48 & 27.12 & 37.44 & 30.40$^\ast$ & 40.10$^\ast$\\
            & Sep-Attack &  8.55 & 23.41 & 12.64 & 26.12 & 28.34 & 39.43 & 91.44$^\ast$ & 95.44$^\ast$\\
            & Co-Attack & 8.79 & 23.74 & 13.10 & 26.07 & 28.79 & 40.03 & 94.76$^\ast$ & 96.89$^\ast$ \\
			& \cellcolor{gray! 20}SGA  & \cellcolor{gray! 20}\textbf{11.42$\pm$0.07} & \cellcolor{gray! 20}\textbf{24.80$\pm$0.28} & \cellcolor{gray! 20}\textbf{14.91$\pm$0.08} & \cellcolor{gray! 20}\textbf{28.82$\pm$0.11} & \cellcolor{gray! 20}\textbf{31.24$\pm$0.42} & \cellcolor{gray! 20}\textbf{42.12$\pm$0.11} & \cellcolor{gray! 20}\textbf{99.24$\pm$0.18$^\ast$} & \cellcolor{gray! 20}\textbf{99.49$\pm$0.05$^\ast$}  \\
			\bottomrule[0.3mm]
	\end{tabular}}
\end{center}
\vspace{-8pt}
\caption{
\textbf{Comparison with state-of-the-art methods on image-text retrieval.}
We report the attack success rate ($\%$) of R@1 for both IR and TR.
The source column indicates the source models used to generate the adversarial data on Flickr30K. $^\ast$ indicates white-box attacks. A higher ASR indicates better adversarial transferability.}
\vspace{-5pt}
\label{tab:t3_main_f30k}
\end{table*}

\subsection{Transferability Analysis}\label{sec: exp_ana}
In this paper, we present a systematic study of the adversarial transferability of VLP models, which has not been explored. 
As demonstrated in Section \ref{sec:analysis}, existing methods, including the separate unimodal adversarial attack (Sep-Attack) and the multimodal adversarial attack (Co-Attack), exhibit limited transferability to other VLP models.

To improve transferability in multimodal learning, we intuitively investigate the adoption of transfer-based attacks from unimodal learning such as image classification. 
Specifically, we consider MI \cite{Dong2018BoostingAA}, DIM \cite{Xie2018ImprovingTransDiverse}, and PNA\_PO \cite{wei2022PNA_PO}. 
However, this approach can be problematic if cross-modal interactions and the unique many-to-many alignments in multimodal learning are not taken into account.

Table \ref{tab:t2_exp_unimodal} illustrates that multimodal attack methods that incorporate transfer-based image attacks exhibit minimal improvement in transferability while compromising white-box performance.
Specifically, when integrated with MI, Co-Attack drops significantly by 12.3\% in white-box settings, while only maintaining 25.40\% ASR in transferability (ALBEF to TCL).
However, our SGA shows superior performance in both white-box and black-box settings.
Notably, Sep-Attack combined with transfer-based attacks not only reduces the effectiveness of white-box attacks but also fails to improve adversarial transferability in almost all black-box settings.
The results provide empirical evidence that directly combining unimodal adversarial attacks in multimodal learning without considering cross-modal interactions and alignments can be problematic, even when using transfer-based unimodal attacks.
Additional discussion is provided in Appendix \ref{sec:supp_b_ana}.
\begin{table}[t]
\begin{center}
\small
\renewcommand\arraystretch{1.1}
\setlength{\tabcolsep}{4pt}
		\scalebox{0.9}[0.9]{
		\begin{tabular}{ @{\extracolsep{\fill}} l|ccccc@{}}
        \toprule
			 {\textbf{Attack}} & {B@4} & {METEOR} & {ROUGE\_L} & {CIDEr} & {SPICE}  \\
   \hline
   \hline
			Baseline & 39.7 & 31.0 & 60.0 & 133.3 & 23.8 \\
          Co-Attack & 37.4  & 29.8 & 58.4 & 125.5 & 22.8 \\
          SGA & \textbf{34.8} & \textbf{28.4} & \textbf{56.3} & \textbf{116.0} & \textbf{21.4} \\
   \bottomrule
	\end{tabular}}
\end{center}
\vspace{-10pt}
\caption{ \textbf{Cross-Task Transferability: ITR $\rightarrow$ IC.}  Adversarial data generated from Image-Text Retrieval (\textbf{ITR}) to attack Image Captioning (\textbf{IC}) on MSCOCO.
The source model and target model are ALBEF and BLIP, respectively.
The Baseline represents the original performance of IC on clean data.
Lower values indicate better adversarial transferability.}
\vspace{-12pt}
\label{tab:t4_exp_caption}
\end{table}

\subsection{Experimental Results}\label{sec: exp_res}
\paragraph{Multimodal Fusion Modules.}
First, we investigate VLP models with different fusion modules, namely, fused VLP models and aligned VLP models. 
We generate adversarial examples on both types of models and evaluate their attack performance when transferred to other VLP models while ensuring consistency in the input size of images. 
For example, adversarial images generated by ALBEF or TCL are resized to $224\times224$ before performing transfer attacks on CLIP, and adversarial examples generated on CLIP are resized to $384\times384$ before being transferred to ALBEF or TCL.

As shown in Table~\ref{tab:t3_main_f30k}, experimental results demonstrate the superiority of our proposed SGA over existing multimodal attack methods in all black-box settings.
Specifically, our SGA achieves significant improvements in adversarial transferability when the source and target models are of the same type. 
For instance, SGA outperforms Co-Attack by approximately 30$\%$ in terms of attack success rate when transferring adversarial data from ALBEF to TCL. 
Moreover, in the more challenging scenario where the source and target models are of different types, SGA also surpasses Co-Attack with higher attack success rates. 
More experiments on the MSCOCO dataset are provided in Appendix \ref{sec:supp_b_main_res}.

\paragraph{Model Architectures.}
Then, we explore VLP models with respect to different model architectures.
Many VLP models commonly use ViTs as the vision encoder, where images are segmented into patches before being processed by the transformer model. 
However, in the case of CLIP, the image encoder consists of two distinct architectures: conventional Convolutional Neural Networks (CNNs) and ViTs.
The transferability between CNNs and ViTs has been well-studied in unimodal learning. Therefore, we also investigate the adversarial transferability of CNN-based and ViT-based CLIP models in multimodal learning.

As shown in Table~\ref{tab:t3_main_f30k}, we observe a similar phenomenon as unimodal learning that compared to CNNs, ViTs show better robustness against adversarial perturbations \cite{Naseer2021IntriguingViT}.
Specifically, for all attack methods, the same adversarial multimodal data have a stronger white-box attack effect on $\rm CLIP_{CNN}$ compared to $\rm CLIP_{ViT}$.
Moreover, the adversarial examples generated on $\rm CLIP_{ViT}$ are found to be more transferable to $\rm CLIP_{CNN}$ than vice versa (38.76\% vs. 31.24\%).

Furthermore, our proposed SGA consistently improves transferability on both CNN-based CLIP and ViT-based CLIP compared to other attacks. 
For instance, SGA increases the adversarial transferability for $\rm CLIP_{ViT}$ by 5.83\% and 6.24\% compared to Co-Attack under the white-box setting and black-box setting, respectively.

\subsection{Cross-Task Transferability}\label{sec:cross_task}
Cross-modal interactions and alignments are the core components of multimodal learning regardless of the task.
Therefore, we conduct extensive experiments to explore the effectiveness of our proposed SGA on two additional V+L tasks: Image Captioning (IC) and Visual Grounding (VG).

\paragraph{Image Captioning.}
Image captioning is a generation-based task, where an input image is encoded into a feature vector and then decoded into a natural language sentence.
In our experiments, we craft adversarial images using the source model (ALBEF) with an image-text retrieval objective and then directly attack the target model (BLIP \cite{Li2022BLIP}) on image captioning.
We employ the MSCOCO dataset, which is suitable for both two tasks, and utilize various evaluation metrics to measure the quality of the generated captions, including BLEU \cite{Papineni2002BLEU}, METEOR \cite{Banerjee2005METEORAA}, ROUGE \cite{Lin2004ROUGEAP}, CIDEr \cite{Vedantam2015CIDEr}, and SPICE \cite{Anderson2016SPICE}.

We present the performance on image captioning of BLIP after being attacked in Table~\ref{tab:t4_exp_caption}. 
Experimental results demonstrate clear improvements in the adversarial transferability of the proposed SGA compared to Co-Attack.
Specifically, our SGA improves the BLEU score by up to 2.6\% and the CIDEr score by up to 9.5\%.

\begin{table}[t]
\begin{center}
\small
\renewcommand\arraystretch{1.1}
\setlength{\tabcolsep}{12pt}
		\scalebox{0.9}[0.9]{
		\begin{tabular}{ @{\extracolsep{\fill}} l|ccc}
        \toprule
			 {\textbf{Attack}} & {Val} & {TestA} & {TestB}    \\
            \hline
            \hline
			Baseline & 58.46 & 65.89 & 46.25  \\
           Co-Attack & 54.26 & 61.80 & 43.81   \\
            SGA & \textbf{53.55} & \textbf{61.19} & \textbf{43.71} \\
   \bottomrule
	\end{tabular}}
\end{center}
\vspace{-8pt}
\caption{ \textbf{Cross-Task Transferability: ITR $\rightarrow$ VG.}  Adversarial data generated from Image-Text Retrieval (\textbf{ITR}) to attack Visual Grounding (\textbf{VG}) on RefCOCO+.
The source model and target model are both ALBEF.
The Baseline represents the original performance of VG on clean data.
Lower values indicate better adversarial transferability.}
\vspace{-10pt}
\label{tab:t5_exp_grounding}
\end{table}

\paragraph{Visual Grounding.}
Visual grounding is another V+L task that aims to localize the region in an image based on the corresponding specific textual description.
Similarly, we generate adversarial images using the source model (ALBEF) from image-text to attack the target model (ALBEF) on visual grounding.
Table~\ref{tab:t5_exp_grounding} shows the results on RefCOCO+ \cite{Yu2016RefCOCO}, where our SGA still outperforms Co-Attack.

\subsection{Ablation Study}\label{sec: exp_ablation}
To systematically investigate the impact of our set-level alignment-preserving augmentations, we conducted ablation experiments on image-text retrieval to evaluate the effect of varying the number of augmented image sets $N$ with multi-scale transformation and the number $M$ of augmented caption sets. 
Specifically, we employed ALBEF as the source model and $\rm CLIP_{ViT}$ as the target model on Flickr30K. 
More details are provided in Appendix \ref{sec:supp_b_ablation}.

\begin{figure}[t]
\begin{center}
   \includegraphics[width=1\linewidth]{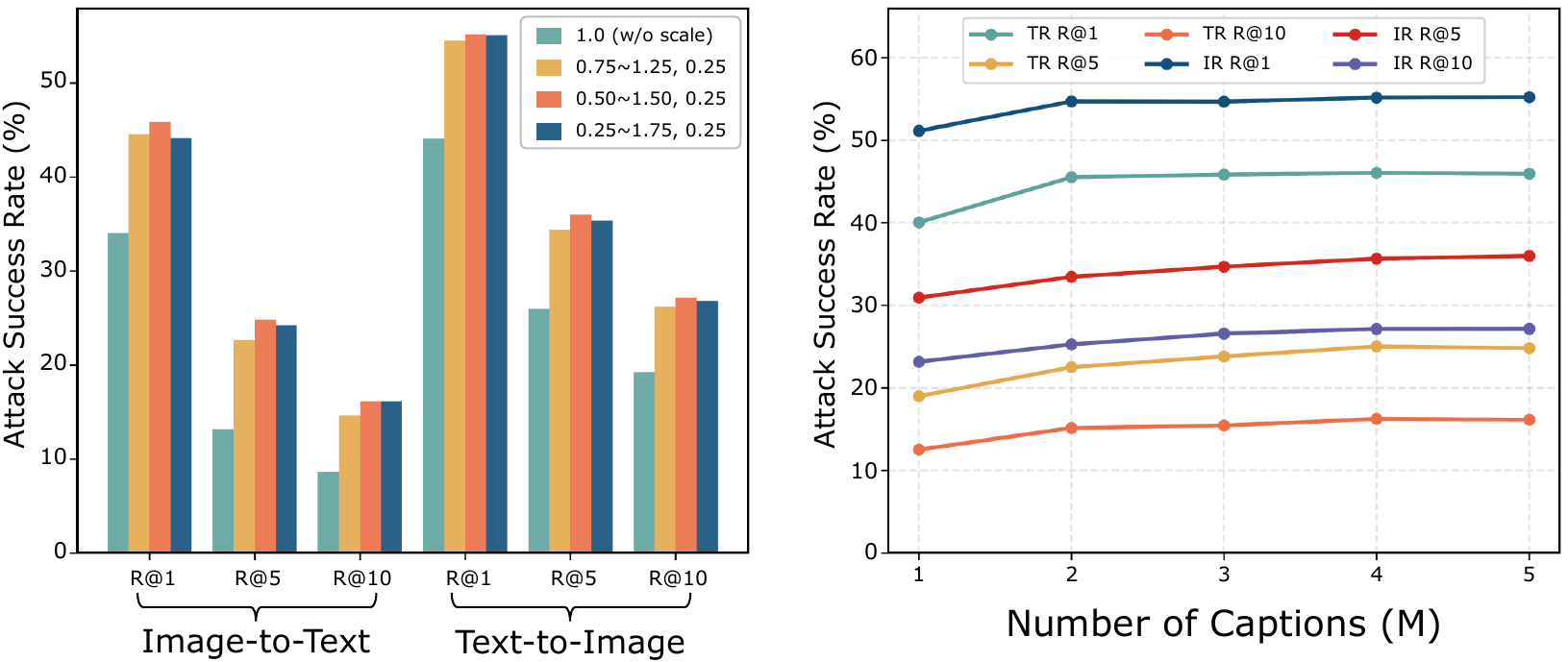}
\end{center}
\vspace{-10pt}
   \caption{\textbf{Ablation Study:} Attack success rates ($\%$) on set-level augmentations, \textbf{Left} for image set and \textbf{Right} for caption set. 
The performance of SGA is consistently improved by using larger augmented image and caption sets.}
\vspace{-10pt}
\label{fig:f5_ablation}
\end{figure}

\paragraph{Multi-scale Image Set.}
We propose the use of multiple scale-invariant images to generate diverse adversarial data in SGA. 
Results in the left of Figure \ref{fig:f5_ablation} reveal that the transferability significantly increases as we introduce more diverse images with different scales, peaking when the scale range is set to $[0.50, 1.50]$ with a step of $0.25$. 
We set the scale range $S=\{0.50, 0.75, 1.00, 1.25, 1.50\}$ for optimal performance.

\paragraph{Multi-pair Caption Set.}
We also conduct experiments to investigate the impact of an enlarged caption set on adversarial transferability. 
The number of additional captions ranged from $1$ to $M$, where $M$ represents the most matching caption pairs from the dataset for each image. 
Results presented in the right panel of Figure \ref{fig:f5_ablation} indicate that if $M > 1$, the black-box performance increases significantly but eventually plateaus.
These results demonstrate the effectiveness of using multiple alignment-preserving inter-modal information to enhance adversarial transferability. 
Furthermore, we observed that the performance is relatively insensitive to the number of extra captions, but adding more captions can improve the overall adversarial transferability.

\section{Conclusion}
\label{sec:conclusion}    
In this paper, we make the first attempt to investigate the adversarial transferability of typical VLP models.
We systematically evaluate the existing attack methods and reveal that they still exhibit lower transferability, despite their impressive performance in white-box settings.
Our investigation highlights the need for specially designed transferable attacks in multimodal learning that can model the many-to-many cross-modal alignments and interactions. 
We propose SGA, a highly transferable multimodal attack, which leverages set-level alignment-preserving augmentations through cross-modal guidance to thoroughly exploit multimodal interactions. 
We hope that this work could inspire further research to evaluate and enhance the adversarial robustness of VLP models.

\paragraph{Acknowledgments.}
This work was supported by the National Key R\&D Program of China (Grant NO. 2022YFF1202903) and the National Natural Science Foundation of China (Grant NO. 62122035).       

{\small
\bibliographystyle{ieee_fullname}
\bibliography{egbib}
}

\clearpage
\appendix
\section{Motivation}
\label{sec:supp_a}

The analysis and discussion presented in Section \ref{sec:analysis} shed light on the keys to improving adversarial transferability among VLP models: multimodal interaction and diverse data.
To further figure out a practicable solution, we delve into the cases of transfer failure of existing attack methods.
We find that half of the failure cases are raised by the existence of multiple paired captions.

Considering an image-text pair $(v,t)$, the corresponding adversarial image $v'$ generated on model $f_{wb}$ in white-box manner (note that only $(v,t)$ and $f_{wb}$ are utilized in the process of crafting $v'$), a black-box model $f_{bb}$ and another several matched captions $\mathbf{t}=\{t_1,...,t_n\}$, we define two events:
\begin{itemize}
    \item \textit{Event A}: adversarial image $v'$ cannot match anyone of the captions $t \cup \{t_1,...,t_n\}$  in model $f_{wb}$. 
    \item \textit{Event B}: adversarial image $v'$ can match one of captions $\{t_1,...,t_n\}$ in model $f_{bb}$.
\end{itemize}
\textit{Event A} indicates that the adversarial image successfully fools model $f_{wb}$, successful case of white-box attack.
\textit{Event B} indicates that the adversarial image cannot fully fool the target model $f_{bb}$ in a transferring manner, failure case of transfer-based black-box attack.
We present the statistic figures of $p($\textit{Event A}$)$ and $p($\textit{Event B} $|$ \textit{Event A}$)$ in Table \ref{tab:supp_t6_black_box_failures}.
As shown in the table, even though the adversarial images have high attack ability in the white-box model (about 71\% - 80\% error rate), around half of them fail due to matching other paired captions when transferring to a black-box model (about 46\%-57\%).

In detail, existing attacks tend to restrict the generated adversarial image $v'$ far away (Euclidean distance or cosine distance in the embedding space in most of the cases) from the original image $v$ or the caption $t$.
These methods only utilize the information of a single image-caption pair $(v,t)$ in their processes of crafting adversarial examples.
As a result, although in most of the cases $v'$ is far away from $t$ and other paired captions $ \{t_1,...,t_n\}$ in the embedding space of the white-box model, it is prone to approaching $ \{t_1,...,t_n\}$ when transferred to a black-box model and the embedding space changes, which means the failure of transfer-based black-box attack.

\begin{figure}[t]
\begin{center}
   \includegraphics[width=0.9\linewidth]{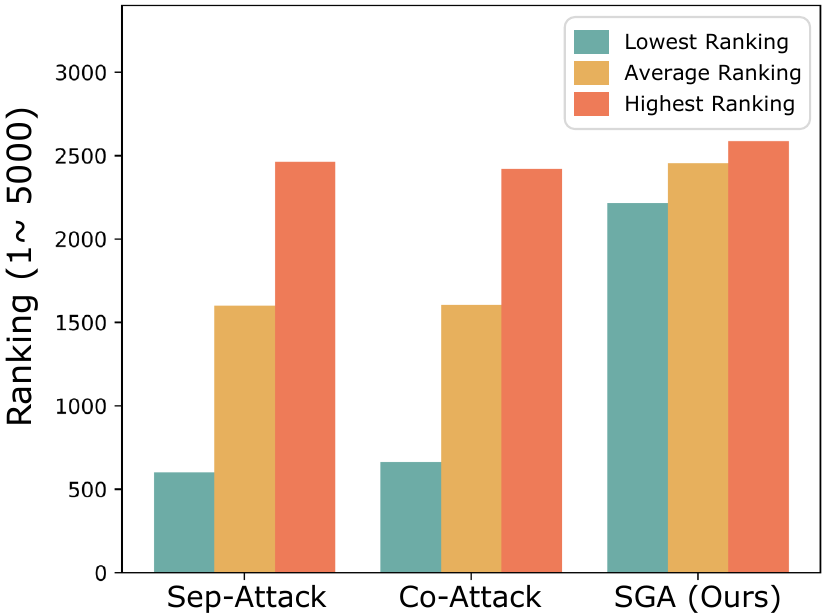}
\end{center}
\vspace{-10pt}
   \caption{The adversarial image may have weak attack ability to some paired captions. The experiment is conducted on model ALBEF, dataset Flickr30K.}
\label{fig:supp_f6_attack_ability}
\end{figure}

We attribute the failure of the transfer attack to the lack of cross-modal interaction (corresponding to the first two rows in Table \ref{tab:supp_t6_black_box_failures}). 
The adversarial image $v'$ generated merely based on image $v$ or single image-text pair $(v,t)$ can have strong attack ability to the caption $t$ and always weak attack ability to $\{t_1,...,t_n\}$.
When transferred to a black-box model, the adversarial image $v'$ may still maintain satisfactory attack ability to caption $t$ but most likely to lose the attack ability to captions $\{t_1,...,t_n\}$.
Note that $v'$ has the attack ability to $t$ in model $f$ means $v'$ cannot successfully match $t$ in the embedding space of model $f$.
To validate the claim, in Figure \ref{fig:supp_f6_attack_ability}, we use the ranking to measure the adversarial image's attack ability to the caption.
Higher ranking, stronger attack ability.
Since an adversarial image has several paired captions in the gallery, we present the lowest, average, and highest ranking of these captions.
As shown in Figure \ref{fig:supp_f6_attack_ability}, for the attack method with no cross-modal interaction (Sep-Attack) and the attack method with single-pair cross-modal interaction (Co-Attack), though the generated adversarial image can have a high attack ability to some captions, there always exists a caption that the adversarial image has weak attack ability to it (the lowest rankings of Sep-attack and Co-Attack are both around 600, which means weak attack ability compared the highest rankings of them, around 2,200 and 2,400).

An implicit assumption in the previous statement is that high attack ability in the white-box model means high adversarial transferability in the black-box model, which can also be verified among existing attack methods.
Considering a image-caption pair $(v,t)$, the corresponding adversarial image $v'$ generated on the white-box model $f_{wb}$ , a black-model $f_{bb}$ and several matched captions of $v$, $\{t_1,...,t_n\}$, we define two events:
\begin{itemize}
    \item \textit{Event C}: adversarial image $v'$ cannot match $t$ in white-box model $f_{wb}$. 
    \item \textit{Event D}: adversarial image $v'$ cannot match $t$ in black-box model $f_{bb}$.
\end{itemize}
We present the statistic figures of $p($\textit{Event C}$)$ and $p($\textit{Event D} $|$ \textit{Event C}$)$ in Table \ref{tab:supp_t7_attack_preservation}.
If the adversarial image $v'$ has a high attack ability to caption $t$ in the white-box model, it is very likely that it also maintains the attack ability towards caption $t$ when transferred to a black-box model.
For example, if the adversarial image $v'$ generated on model ALBEF succeeds in attacking caption $t$ in model ALBEF, there is a high probability that it can succeed in attacking caption $t$ in model TCL, 40.51\%, compared to the overall adversarial transferability from ALBEF to TCL, 15.21\%.

According to the analysis above, to boost the adversarial transferability of the generated adversarial image, it is crucial to consider multiple paired captions and push the adversarial image away from all the paired captions, thus preserving the attack ability when transferring to other black-box models.
Crafting adversarial captions for high transferability follows a similar approach, which can also benefit from more paired images.

\begin{table}[t]
\begin{center}
\footnotesize
\renewcommand\arraystretch{1}
\setlength{\tabcolsep}{2.2pt}
		\scalebox{0.97}[0.97]{
		\begin{tabular}{  l|c|c|ccc}
        \toprule[0.3mm]
			&  &{$p$ (\textit{Event A})} $\uparrow$ & \multicolumn{3}{c}{$p$ (\textit{Event B} $|$\textit{ Event A}) $\downarrow$} \\
			\cmidrule{3-6}
			\multirow{-2}{*}{\textbf{Attack}} & \multirow{-2}{*}{\makecell[c]{Cross-modal \\ Interaction}}& {ALBEF} & {TCL} & {CLIP$_{\rm ViT}$} & {CLIP$_{\rm CNN}$}\\
   
			\midrule
            Sep-Attack  &  No   & 74.60\%    & 46.78\%   & 41.69\%   & 41.82\%   \\
            Co-Attack   &  Single-pair  & 80.60\%    & 50.50\%   & 40.82\%   & 42.93\%   \\
            SGA (Ours)         &  Set-level     & \textbf{97.20\%}    & \textbf{28.91\%}  & \textbf{34.67\%}   & \textbf{38.58\%}\\
			\bottomrule[0.3mm]
	\end{tabular}}
\end{center}
\vspace{-8pt}
\caption{ 
\textit{\textbf{Event A}}: adversarial image $v'$ cannot match anyone of $t \cup \{t_1,...,t_n\}$ in white-box model $f_{wb}$.
\textit{\textbf{Event B}}: adversarial image $v'$ can match one of $\{t_1,...,t_n\}$ in black-box model $f_{bb}$.
The adversarial data is generated on model ALBEF, dataset Flickr30K.
}
\label{tab:supp_t6_black_box_failures}
\end{table}

\begin{table}[t]
\begin{center}
\footnotesize
\renewcommand\arraystretch{1}
\setlength{\tabcolsep}{2.2pt}
		\scalebox{0.97}[0.97]{
		\begin{tabular}{  l|c|c|ccc}
        \toprule[0.3mm]
			&  &{$p$ (\textit{Event C})} $\uparrow$ & \multicolumn{3}{c}{$p$ (\textit{Event D} $|$\textit{ Event C}) $\uparrow$} \\
			\cmidrule{3-6}
			\multirow{-2}{*}{\textbf{Attack}} & \multirow{-2}{*}{\makecell[c]{Cross-modal \\ Interaction}}& {ALBEF} & {TCL} & {CLIP$_{\rm ViT}$} & {CLIP$_{\rm CNN}$}\\
   
			\midrule
            Sep-Attack  &  No   & 83.30\%    & 33.37\%   & 38.77\%   & 45.38\%   \\
            Co-Attack   &  Single-pair  & 89.60\%    & 40.51\%   & 39.62\%   & 47.32\%   \\
            SGA (Ours)         &  Set-level     & \textbf{98.20\%}    & \textbf{54.38\%}  & \textbf{46.84\%}   & \textbf{55.61\%}\\
			\bottomrule[0.3mm]
	\end{tabular}}
\end{center}
\vspace{-8pt}
\caption{ 
\textit{\textbf{Event C}}: adversarial image $v'$ cannot match caption $t$ in white-box model.
\textit{\textbf{Event D}}: adversarial image $v'$ cannot match caption $t$ in black-box model.
The adversarial data is generated on model ALBEF, dataset Flickr30K.
}
\vspace{-10pt}
\label{tab:supp_t7_attack_preservation}
\end{table}

\section{Experiments \& Analysis}
\label{sec:supp_b}

\subsection{Experimental Settings} 
\label{sec:supp_b_exp_setting}
Since fused VLP models contain both multimodal encoder and unimodal encoder, two types of embedding can be perturbed, \textit{i.e.}, multimodal embedding, and unimodal embedding. 
The embeddings can be further divided into the full embedding (denoted as {\texttt{Multi$\rm_{full}$}} or {\texttt{Uni$\rm_{full}$}}) and [CLS] of embedding (denoted as {\texttt{Multi$\rm_{CLS}$}} or {\texttt{Uni$\rm_{CLS}$}}).
For aligned VLP models (\textit{e.g.}, CLIP \cite{Radford2021CLIP}), since the image encoder can be ViT or CNN, only [CLS] of embedding for CLIP$_{\rm ViT}$  is discussed and consider the embedding of CLIP$_{\rm CNN}$ as [CLS] of embedding \cite{Zhang2022Co-attack}.

\begin{table*}[t]
\begin{center}
\small
\renewcommand\arraystretch{1}
  \setlength{\tabcolsep}{3mm}{
		\begin{tabular}{l|l|l|ccc|ccc}
        \toprule
        \multicolumn{9}{c}{\textbf{{\textbf{\fontsize{10.5pt}{\baselineskip}\selectfont{Sep-Attack}}}}} \\
        \midrule
        \multirow{2}{*}{\textbf{\fontsize{11pt}{\baselineskip}\selectfont{Source}}} &
        \multirow{2}{*}{\textbf{\fontsize{11pt}{\baselineskip}\selectfont{Attack}}} & \multirow{2}{*}{\textbf{\fontsize{11pt}{\baselineskip}\selectfont{Target}}} & \multicolumn{3}{c|}{\textbf{\fontsize{10.5pt}{\baselineskip}\selectfont{Image-to-Text}}} & \multicolumn{3}{c}{\textbf{\fontsize{10.5pt}{\baselineskip}\selectfont{Text-to-Image}}}   \\
                    &  &   & R@1  & R@5  & R@10    & R@1  & R@5  & R@10 \\
      \hline
      \multirow{48}{*}{\textbf{ALBEF}}                      & \multirow{4}{*}{\texttt{\textbf{Text}@Uni$\rm_{full}$}}       & ALBEF  & \textbf{8.34$^\ast$} & \textbf{1.40$^\ast$} & \textbf{0.60$^\ast$} & \textbf{21.19$^\ast$} & \textbf{11.36$^\ast$} & \textbf{9.18$^\ast$}   \\
&  & TCL  & 7.90 & 1.21 & 0.30 & 19.45 & 8.87 & 6.26   \\
&  & CLIP$_{\rm ViT}$  & 23.31 & 9.55 & 4.88 & 36.05 & 20.77 & 15.98   \\
&  & CLIP$_{\rm CNN}$  & 26.05 & 9.73 & 5.97 & 38.04 & 21.83 & 16.85   \\
      \cline{2-9}
      & \multirow{4}{*}{\texttt{\textbf{Image}@Uni$\rm_{full}$}}      & ALBEF  & \textbf{62.46$^\ast$} & \textbf{50.70$^\ast$} & \textbf{45.00$^\ast$} & \textbf{68.73$^\ast$} & \textbf{57.38$^\ast$} & \textbf{52.12$^\ast$}   \\
&  & TCL  & 5.48 & 1.21 & 0.80 & 10.43 & 3.33 & 1.89   \\
&  & CLIP$_{\rm ViT}$  & 7.36 & 1.66 & 0.61 & 13.18 & 5.21 & 3.10   \\
&  & CLIP$_{\rm CNN}$  & 10.09 & 2.85 & 1.24 & 15.54 & 6.28 & 3.61   \\
      \cline{2-9}
      & \multirow{4}{*}{\texttt{\textbf{Bi}@Uni$\rm_{full}$}}     & ALBEF  & \textbf{68.93$^\ast$} & \textbf{55.21$^\ast$} & \textbf{49.40$^\ast$} & \textbf{76.33$^\ast$} & \textbf{65.46$^\ast$} & \textbf{59.66$^\ast$}   \\
&  & TCL  & 16.86 & 3.32 & 1.70 & 27.07 & 13.27 & 8.79   \\
&  & CLIP$_{\rm ViT}$  & 25.40 & 9.55 & 4.88 & 36.15 & 20.93 & 15.57   \\
&  & CLIP$_{\rm CNN}$  & 26.82 & 9.73 & 6.49 & 38.80 & 22.34 & 16.90   \\
        \cline{2-9}
       & \multirow{4}{*}{\texttt{\textbf{Text}@Multi$\rm_{full}$}}       & ALBEF  & \textbf{15.43$^\ast$} & \textbf{2.91$^\ast$} & \textbf{1.40$^\ast$} & \textbf{30.54$^\ast$} & \textbf{16.41$^\ast$} & \textbf{12.66$^\ast$}   \\
&  & TCL  & 12.64 & 2.21 & 0.60 & 28.64 & 14.62 & 10.40   \\
&  & CLIP$_{\rm ViT}$  & 26.75 & 10.49 & 5.28 & 41.33 & 24.62 & 19.15   \\
&  & CLIP$_{\rm CNN}$  & 30.27 & 12.16 & 7.11 & 43.43 & 26.64 & 20.96   \\
      \cline{2-9}
      & \multirow{4}{*}{\texttt{\textbf{Image}@Multi$\rm_{full}$}}       & ALBEF  & \textbf{35.97$^\ast$} & \textbf{25.35$^\ast$} & \textbf{21.40$^\ast$} & \textbf{50.54$^\ast$} & \textbf{40.57$^\ast$} & \textbf{37.24$^\ast$}   \\
&  & TCL  & 1.79 & 0.50 & 0.20 & 6.50 & 1.88 & 1.10   \\
&  & CLIP$_{\rm ViT}$  & 7.12 & 1.56 & 0.30 & 13.02 & 5.05 & 3.03   \\
&  & CLIP$_{\rm CNN}$  & 9.83 & 2.85 & 1.34 & 14.75 & 5.56 & 3.23   \\
      \cline{2-9}
      & \multirow{4}{*}{\texttt{\textbf{Bi}@Multi$\rm_{full}$}}       & ALBEF  & \textbf{51.09$^\ast$} & \textbf{36.97$^\ast$} & \textbf{31.90$^\ast$} & \textbf{64.17$^\ast$} & \textbf{52.87$^\ast$} & \textbf{48.73$^\ast$}   \\
&  & TCL  & 16.86 & 4.02 & 1.30 & 32.57 & 16.77 & 12.08   \\
&  & CLIP$_{\rm ViT}$  & 27.48 & 10.70 & 5.69 & 41.78 & 25.02 & 18.88   \\
&  & CLIP$_{\rm CNN}$  & 31.16 & 12.16 & 6.90 & 43.77 & 26.56 & 21.12   \\
        \cline{2-9}
      & \multirow{4}{*}{\texttt{\textbf{Text}@Uni$\rm_{CLS}$}}     & ALBEF  & \textbf{11.57$^\ast$} & \textbf{1.80$^\ast$} & \textbf{1.10$^\ast$} & \textbf{27.46$^\ast$} & \textbf{14.48$^\ast$} & \textbf{10.98$^\ast$}   \\
&  & TCL  & 12.64 & 2.51 & 0.90 & 28.07 & 14.39 & 10.26   \\
&  & CLIP$_{\rm ViT}$  & 29.33 & 11.63 & 6.30 & 43.17 & 26.37 & 19.91   \\
&  & CLIP$_{\rm CNN}$  & 32.69 & 15.43 & 8.65 & 46.11 & 28.43 & 22.14   \\
      \cline{2-9}
      & \multirow{4}{*}{\texttt{\textbf{Image}@Uni$\rm_{CLS}$}}      & ALBEF  & \textbf{52.45$^\ast$} & \textbf{36.57$^\ast$} & \textbf{30.00$^\ast$} & \textbf{58.65$^\ast$} & \textbf{44.85$^\ast$} & \textbf{38.98$^\ast$}   \\
&  & TCL  & 3.06 & 0.40 & 0.10 & 6.79 & 2.21 & 1.20   \\
&  & CLIP$_{\rm ViT}$  & 8.96 & 1.66 & 0.41 & 13.21 & 5.19 & 3.05   \\
&  & CLIP$_{\rm CNN}$  & 10.34 & 2.96 & 1.85 & 14.65 & 5.60 & 3.39   \\
      \cline{2-9}
      & \multirow{4}{*}{\texttt{\textbf{Bi}@Uni$\rm_{CLS}$}}      & ALBEF  & \textbf{65.69$^\ast$} & \textbf{47.60$^\ast$} & \textbf{42.10$^\ast$} & \textbf{73.95$^\ast$} & \textbf{59.50$^\ast$} & \textbf{53.70$^\ast$}   \\
&  & TCL  & 17.60 & 3.72 & 1.90 & 32.95 & 17.10 & 11.90   \\
&  & CLIP$_{\rm ViT}$  & 31.17 & 12.05 & 7.01 & 45.23 & 25.93 & 19.95   \\
&  & CLIP$_{\rm CNN}$  & 32.82 & 15.86 & 9.06 & 45.49 & 28.43 & 22.32   \\
    \cline{2-9}
        & \multirow{4}{*}{\texttt{\textbf{Text}@Multi$\rm_{CLS}$}}       & ALBEF  & \textbf{15.43$^\ast$} & \textbf{2.81$^\ast$} & \textbf{1.30$^\ast$} & \textbf{30.47$^\ast$} & \textbf{15.85$^\ast$} & \textbf{11.85$^\ast$}   \\
&  & TCL  & 13.59 & 3.02 & 1.20 & 30.26 & 15.42 & 11.09   \\
&  & CLIP$_{\rm ViT}$  & 27.12 & 11.94 & 6.50 & 42.53 & 25.20 & 19.36   \\
&  & CLIP$_{\rm CNN}$  & 30.78 & 13.21 & 7.52 & 44.39 & 28.07 & 21.89   \\
      \cline{2-9}
      & \multirow{4}{*}{\texttt{\textbf{Image}@Multi$\rm_{CLS}$}}       & ALBEF  & \textbf{30.76$^\ast$} & \textbf{21.24$^\ast$} & \textbf{17.10$^\ast$} & \textbf{43.85$^\ast$} & \textbf{34.84$^\ast$} & \textbf{31.44$^\ast$}   \\
&  & TCL  & 2.53 & 0.20 & 0.00 & 6.74 & 1.98 & 1.20   \\
&  & CLIP$_{\rm ViT}$  & 7.98 & 1.35 & 0.30 & 12.85 & 5.00 & 3.16   \\
&  & CLIP$_{\rm CNN}$  & 9.96 & 2.64 & 1.75 & 14.92 & 5.65 & 3.37   \\
      \cline{2-9}
      & \multirow{4}{*}{\texttt{\textbf{Bi}@Multi$\rm_{CLS}$}}      & ALBEF  & \textbf{42.13$^\ast$} & \textbf{26.95$^\ast$} & \textbf{22.20$^\ast$} & \textbf{57.76$^\ast$} & \textbf{44.91$^\ast$} & \textbf{39.95$^\ast$}   \\
&  & TCL  & 16.65 & 3.92 & 1.90 & 34.02 & 17.16 & 12.10   \\
&  & CLIP$_{\rm ViT}$  & 28.71 & 11.42 & 6.30 & 42.01 & 24.90 & 18.99   \\
&  & CLIP$_{\rm CNN}$  & 31.03 & 14.16 & 8.96 & 43.98 & 27.17 & 21.30   \\
        \bottomrule[0.3mm]
    \end{tabular}}
\end{center}
\caption{
\textbf{Attack success rates ($\%$)} with different adversarial input modalities under Sep-Attack on image-text retrieval. The adversaries are crafted on ALBEF using Flickr30K. $^\ast$ indicates white-box attacks. A higher ASR indicates better adversarial transferability.}
\label{tab:supp_t8_exp_sep_attack_albef}
\end{table*}

\begin{table*}[t]
\begin{center}
\small
\renewcommand\arraystretch{1}
  \setlength{\tabcolsep}{3mm}{
\begin{tabular}{l|l|l|ccc|ccc}
        \toprule
        \multicolumn{9}{c}{\textbf{{\textbf{\fontsize{10.5pt}{\baselineskip}\selectfont{Sep-Attack}}}}} \\
        \midrule
        \multirow{2}{*}{\textbf{\fontsize{11pt}{\baselineskip}\selectfont{Source}}} &
        \multirow{2}{*}{\textbf{\fontsize{11pt}{\baselineskip}\selectfont{Attack}}} & \multirow{2}{*}{\textbf{\fontsize{11pt}{\baselineskip}\selectfont{Target}}} & \multicolumn{3}{c|}{\textbf{\fontsize{10.5pt}{\baselineskip}\selectfont{Image-to-Text}}} & \multicolumn{3}{c}{\textbf{\fontsize{10.5pt}{\baselineskip}\selectfont{Text-to-Image}}}   \\
                    &  &   & R@1  & R@5  & R@10    & R@1  & R@5  & R@10 \\
      \hline
      \multirow{48}{*}{\textbf{TCL}}                      & \multirow{4}{*}{\texttt{\textbf{Text}@Uni$\rm_{full}$}}       & TCL  & \textbf{9.48$^\ast$} & \textbf{1.51$^\ast$} & \textbf{0.60$^\ast$} & \textbf{23.50$^\ast$} & \textbf{11.83$^\ast$} & \textbf{8.53$^\ast$}   \\
&  & ALBEF  & 9.91 & 1.80 & 0.70 & 23.64 & 12.80 & 10.07   \\
&  & CLIP$_{\rm ViT}$  & 25.89 & 8.41 & 4.57 & 39.79 & 24.22 & 18.62   \\
&  & CLIP$_{\rm CNN}$  & 28.35 & 11.21 & 7.11 & 41.96 & 26.30 & 20.42   \\
      \cline{2-9}
      & \multirow{4}{*}{\texttt{\textbf{Image}@Uni$\rm_{full}$}}      & TCL  & \textbf{45.10$^\ast$} & \textbf{34.07$^\ast$} & \textbf{28.76$^\ast$} & \textbf{53.21$^\ast$} & \textbf{38.27$^\ast$} & \textbf{32.49$^\ast$}   \\
&  & ALBEF  & 3.86 & 0.90 & 0.20 & 7.62 & 2.40 & 1.29   \\
&  & CLIP$_{\rm ViT}$  & 7.12 & 1.66 & 0.71 & 12.82 & 5.35 & 3.14   \\
&  & CLIP$_{\rm CNN}$  & 9.07 & 2.54 & 1.75 & 15.68 & 5.70 & 3.39   \\
      \cline{2-9}
      & \multirow{4}{*}{\texttt{\textbf{Bi}@Uni$\rm_{full}$}}      & TCL  & \textbf{55.95$^\ast$} & \textbf{39.80$^\ast$} & \textbf{33.77$^\ast$} & \textbf{65.38$^\ast$} & \textbf{49.58$^\ast$} & \textbf{42.28$^\ast$}   \\
&  & ALBEF  & 15.02 & 4.01 & 2.60 & 30.47 & 17.04 & 13.28   \\
&  & CLIP$_{\rm ViT}$  & 27.48 & 8.10 & 4.37 & 39.85 & 24.50 & 18.23   \\
&  & CLIP$_{\rm CNN}$  & 29.50 & 11.21 & 7.52 & 42.13 & 26.47 & 20.53   \\
        \cline{2-9}
       & \multirow{4}{*}{\texttt{\textbf{Text}@Multi$\rm_{full}$}}       & TCL  & \textbf{12.86$^\ast$} & \textbf{2.81$^\ast$} & \textbf{1.00$^\ast$} & \textbf{30.33$^\ast$} & \textbf{15.32$^\ast$} & \textbf{10.89$^\ast$}   \\
&  & ALBEF  & 13.24 & 2.61 & 1.20 & 27.13 & 15.16 & 11.28   \\
&  & CLIP$_{\rm ViT}$  & 26.75 & 9.24 & 4.57 & 40.85 & 24.57 & 18.77   \\
&  & CLIP$_{\rm CNN}$  & 28.35 & 11.31 & 6.49 & 42.95 & 26.13 & 20.71   \\
      \cline{2-9}
      & \multirow{4}{*}{\texttt{\textbf{Image}@Multi$\rm_{full}$}}       & TCL  & \textbf{52.05$^\ast$} & \textbf{41.81$^\ast$} & \textbf{35.47$^\ast$} & \textbf{63.05$^\ast$} & \textbf{51.46$^\ast$} & \textbf{46.67$^\ast$}   \\
&  & ALBEF  & 4.38 & 1.50 & 0.90 & 9.87 & 3.28 & 2.10   \\
&  & CLIP$_{\rm ViT}$  & 7.73 & 1.97 & 0.41 & 13.56 & 5.68 & 3.34   \\
&  & CLIP$_{\rm CNN}$  & 9.32 & 2.64 & 1.44 & 14.92 & 5.70 & 3.32   \\
      \cline{2-9}
      & \multirow{4}{*}{\texttt{\textbf{Bi}@Multi$\rm_{full}$}}       & TCL  & \textbf{61.96$^\ast$} & \textbf{48.64$^\ast$} & \textbf{42.08$^\ast$} & \textbf{71.74$^\ast$} & \textbf{61.07$^\ast$} & \textbf{55.83$^\ast$}   \\
&  & ALBEF  & 19.29 & 6.21 & 3.00 & 35.17 & 19.71 & 14.96   \\
&  & CLIP$_{\rm ViT}$  & 26.75 & 9.55 & 5.08 & 41.37 & 24.78 & 18.71   \\
&  & CLIP$_{\rm CNN}$  & 30.78 & 11.31 & 7.42 & 43.53 & 26.23 & 20.42   \\
        \cline{2-9}
      & \multirow{4}{*}{\texttt{\textbf{Text}@Uni$\rm_{CLS}$}}     
      & TCL  & \textbf{14.54$^\ast$} & \textbf{2.31$^\ast$} & \textbf{0.60$^\ast$} & \textbf{29.17$^\ast$} & \textbf{15.03$^\ast$} & \textbf{10.91$^\ast$}   \\
&  & ALBEF  & 11.89 & 2.20 & 0.70 & 26.82 & 14.09 & 10.80   \\
&  & CLIP$_{\rm ViT}$  & 29.69 & 12.77 & 7.62 & 44.49 & 27.47 & 21.00   \\
&  & CLIP$_{\rm CNN}$  & 33.46 & 14.38 & 9.37 & 46.07 & 29.28 & 22.59   \\
      \cline{2-9}
      & \multirow{4}{*}{\texttt{\textbf{Image}@Uni$\rm_{CLS}$}}      & TCL  & \textbf{77.87$^\ast$} & \textbf{65.13$^\ast$} & \textbf{58.72$^\ast$} & \textbf{79.48$^\ast$} & \textbf{66.26$^\ast$} & \textbf{60.36$^\ast$}   \\
&  & ALBEF  & 6.15 & 1.30 & 0.70 & 10.78 & 3.36 & 1.70   \\
&  & CLIP$_{\rm ViT}$  & 7.48 & 1.45 & 0.81 & 13.72 & 5.37 & 3.01   \\
&  & CLIP$_{\rm CNN}$  & 10.34 & 2.75 & 1.54 & 15.33 & 5.77 & 3.28   \\
      \cline{2-9}
      & \multirow{4}{*}{\texttt{\textbf{Bi}@Uni$\rm_{CLS}$}}      & TCL  & \textbf{84.72$^\ast$} & \textbf{73.07$^\ast$} & \textbf{65.43$^\ast$} & \textbf{86.07$^\ast$} & \textbf{74.67$^\ast$} & \textbf{68.83$^\ast$}   \\
&  & ALBEF  & 20.13 & 4.91 & 2.70 & 36.48 & 19.48 & 14.82   \\
&  & CLIP$_{\rm ViT}$  & 31.29 & 12.98 & 7.72 & 44.65 & 26.82 & 20.37   \\
&  & CLIP$_{\rm CNN}$  & 33.33 & 14.27 & 9.89 & 45.80 & 29.18 & 23.02   \\
    \cline{2-9}
        & \multirow{4}{*}{\texttt{\textbf{Text}@Multi$\rm_{CLS}$}}       & TCL  & \textbf{18.34$^\ast$} & \textbf{4.02$^\ast$} & \textbf{1.90$^\ast$} & \textbf{33.90$^\ast$} & \textbf{16.68$^\ast$} & \textbf{11.78$^\ast$}   \\
&  & ALBEF  & 13.66 & 2.30 & 0.90 & 27.90 & 14.11 & 10.31   \\
&  & CLIP$_{\rm ViT}$  & 27.85 & 11.32 & 6.71 & 42.01 & 24.95 & 18.88   \\
&  & CLIP$_{\rm CNN}$  & 30.27 & 13.95 & 8.34 & 44.32 & 27.58 & 21.23   \\
      \cline{2-9}
      & \multirow{4}{*}{\texttt{\textbf{Image}@Multi$\rm_{CLS}$}}       & TCL  & \textbf{37.41$^\ast$} & \textbf{28.04$^\ast$} & \textbf{24.15$^\ast$} & \textbf{48.93$^\ast$} & \textbf{39.01$^\ast$} & \textbf{35.72$^\ast$}   \\
&  & ALBEF  & 2.92 & 0.90 & 0.50 & 8.07 & 2.65 & 1.62   \\
&  & CLIP$_{\rm ViT}$  & 7.48 & 1.77 & 0.41 & 13.34 & 4.91 & 3.10   \\
&  & CLIP$_{\rm CNN}$  & 9.58 & 3.07 & 1.54 & 15.40 & 5.26 & 3.25   \\
      \cline{2-9}
      & \multirow{4}{*}{\texttt{\textbf{Bi}@Multi$\rm_{CLS}$}}      & TCL  & \textbf{47.31$^\ast$} & \textbf{34.77$^\ast$} & \textbf{29.66$^\ast$} & \textbf{60.31$^\ast$} & \textbf{48.07$^\ast$} & \textbf{43.36$^\ast$}   \\
&  & ALBEF  & 18.87 & 5.21 & 2.70 & 34.03 & 18.17 & 13.12   \\
&  & CLIP$_{\rm ViT}$  & 28.47 & 11.63 & 6.30 & 42.53 & 25.53 & 19.06   \\
&  & CLIP$_{\rm CNN}$  & 31.03 & 14.06 & 8.55 & 44.46 & 27.00 & 20.74   \\
        \bottomrule[0.3mm]
    \end{tabular}}
\end{center}
\caption{
\textbf{Attack success rates ($\%$)} with different adversarial input modalities under Sep-Attack on image-text retrieval. The adversaries are crafted on TCL using Flickr30K. $^\ast$ indicates white-box attacks. A higher ASR indicates better adversarial transferability.}
\label{tab:supp_t9_exp_sep_attack_tcl}
\end{table*}

\begin{table*}[t]
\begin{center}
\small
\renewcommand\arraystretch{1}
  \setlength{\tabcolsep}{3mm}{
\begin{tabular}{l|l|l|ccc|ccc}
        \toprule
        \multicolumn{9}{c}{\textbf{{\textbf{\fontsize{10.5pt}{\baselineskip}\selectfont{Sep-Attack}}}}} \\
        \midrule
        \multirow{2}{*}{\textbf{\fontsize{10pt}{\baselineskip}\selectfont{Source}}} &
        \multirow{2}{*}{\textbf{\fontsize{10pt}{\baselineskip}\selectfont{Attack}}} & \multirow{2}{*}{\textbf{\fontsize{10pt}{\baselineskip}\selectfont{Target}}} & \multicolumn{3}{c|}{\textbf{\fontsize{10pt}{\baselineskip}\selectfont{Image-to-Text}}} & \multicolumn{3}{c}{\textbf{\fontsize{10pt}{\baselineskip}\selectfont{Text-to-Image}}}   \\
                    &  &   & R@1  & R@5  & R@10    & R@1  & R@5  & R@10 \\
      \hline
      \multirow{12}{*}{\textbf{CLIP$_{\rm ViT}$}}                      & \multirow{4}{*}{\texttt{\textbf{Text}@Uni}}       & CLIP$_{\rm ViT}$   & \textbf{28.34$^\ast$} & \textbf{11.73$^\ast$} & \textbf{6.81$^\ast$} & \textbf{39.08$^\ast$} & \textbf{24.08$^\ast$} & \textbf{17.44$^\ast$} \\
      &     & CLIP$_{\rm CNN}$  & 30.40 & 11.63 & 5.97 & 37.43 & 24.96 & 18.66  \\
      &     & ALBEF  & 9.59 & 1.30 & 0.40 & 22.64 & 10.95 & 8.17 \\
      &     & TCL    & 11.80 & 1.91 & 0.70 & 25.07 & 12.92 & 8.90  \\
      \cline{2-9}
      & \multirow{4}{*}{\texttt{\textbf{Image}@Uni}}       & CLIP$_{\rm ViT}$   & \textbf{70.92$^\ast$} & \textbf{50.05$^\ast$} & \textbf{42.28$^\ast$} & \textbf{78.61$^\ast$} & \textbf{60.78$^\ast$} & \textbf{51.50$^\ast$}  \\
      &     & CLIP$_{\rm CNN}$  & 5.36 & 1.16 & 0.72 & 8.44 & 2.35 & 1.54 \\
      &     & ALBEF  & 2.50 & 0.40 & 0.10 & 4.93 & 1.44 & 1.01  \\
      &     & TCL    & 4.85 & 0.20 & 0.20 & 8.17 & 2.27 & 1.46  \\
      \cline{2-9}
      & \multirow{4}{*}{\texttt{\textbf{Bi}@Uni}}       & CLIP$_{\rm ViT}$   & \textbf{79.75$^\ast$} & \textbf{63.03$^\ast$} & \textbf{53.76$^\ast$} & \textbf{86.79$^\ast$} & \textbf{75.24$^\ast$} & \textbf{67.84$^\ast$} \\
      &     & CLIP$_{\rm CNN}$  & 30.78 & 12.16 & 6.39 & 39.76 & 25.62 & 19.34 \\
      &     & ALBEF  & 9.59 & 1.30 & 0.50 & 23.25 & 11.22 & 8.01  \\
      &     & TCL    & 11.38 & 2.11 & 0.90 & 25.60 & 12.92 & 9.14 \\
        \hline
      \multirow{12}{*}{\textbf{CLIP$_{\rm CNN}$}}                      & \multirow{4}{*}{\texttt{\textbf{Text}@Uni}}       & CLIP$_{\rm CNN}$   & \textbf{30.40$^\ast$} & \textbf{13.00$^\ast$} & \textbf{7.31$^\ast$} & \textbf{40.10$^\ast$} & \textbf{26.71$^\ast$} & \textbf{20.85$^\ast$} \\
      &     & CLIP$_{\rm ViT}$  & 27.12 & 11.21 & 6.81 & 37.44 & 23.48 & 17.66 \\
      &     & ALBEF  & 8.86 & 1.50 & 0.60 & 23.27 & 11.34 & 8.41 \\
      &     & TCL    & 12.33 & 2.01 & 0.90 & 25.48 & 13.25 & 8.81  \\
      \cline{2-9}
      & \multirow{4}{*}{\texttt{\textbf{Image}@Uni}}       & CLIP$_{\rm CNN}$   & \textbf{86.46$^\ast$} & \textbf{69.13$^\ast$} & \textbf{61.17$^\ast$} & \textbf{92.25$^\ast$} & \textbf{81.00$^\ast$} & \textbf{75.04$^\ast$} \\
      &     & CLIP$_{\rm ViT}$  & 1.10 & 0.52 & 0.41 & 6.60 & 2.73 & 1.48 \\
      &     & ALBEF  & 2.09 & 0.30 & 0.10 & 4.82 & 1.29 & 0.87  \\
      &     & TCL    & 4.00 & 0.40 & 0.20 & 7.81 & 2.09 & 1.34 \\
      \cline{2-9}
      & \multirow{4}{*}{\texttt{\textbf{Bi}@Uni}}       & CLIP$_{\rm CNN}$   & \textbf{91.44$^\ast$} & \textbf{78.54$^\ast$} & \textbf{71.58$^\ast$} & \textbf{95.44$^\ast$} & \textbf{88.48$^\ast$} & \textbf{82.88$^\ast$} \\
      &     & CLIP$_{\rm ViT}$  & 28.34 & 10.8 & 6.30 & 39.43 & 24.34 & 18.36 \\
      &     & ALBEF  & 8.55 & 1.50 & 0.60 & 23.41 & 11.38 & 8.23  \\
      &     & TCL    & 12.64 & 1.91 & 0.70 & 26.12 & 13.44 & 8.96 \\
        \bottomrule[0.3mm]
    \end{tabular}}
\end{center}
\caption{
\textbf{Attack success rates ($\%$)} with different adversarial input modalities under Sep-Attack on image-text retrieval. The adversaries are crafted on CLIP using Flickr30K. $^\ast$ indicates white-box attacks. A higher ASR indicates better adversarial transferability.}
\label{tab:supp_t10_exp_sep_attack_clip}
\end{table*}

\begin{table*}[t]
\begin{center}
\small
\renewcommand\arraystretch{1}
  \setlength{\tabcolsep}{3mm}{
\begin{tabular}{l|l|l|ccc|ccc}
        \toprule
        \multicolumn{9}{c}{\textbf{{\textbf{\fontsize{10.5pt}{\baselineskip}\selectfont{Co-Attack}}}}} \\
        \midrule
        \multirow{2}{*}{\textbf{\fontsize{11pt}{\baselineskip}\selectfont{Source}}} &
        \multirow{2}{*}{\textbf{\fontsize{11pt}{\baselineskip}\selectfont{Attack}}} & \multirow{2}{*}{\textbf{\fontsize{11pt}{\baselineskip}\selectfont{Target}}} & \multicolumn{3}{c|}{\textbf{\fontsize{10.5pt}{\baselineskip}\selectfont{Image-to-Text}}} & \multicolumn{3}{c}{\textbf{\fontsize{10.5pt}{\baselineskip}\selectfont{Text-to-Image}}}   \\
                    &  &   & R@1  & R@5  & R@10    & R@1  & R@5  & R@10 \\
      \hline
      \multirow{12}{*}{\textbf{ALBEF}}                      & \multirow{4}{*}{\texttt{\textbf{Text}@Multi}}       & ALBEF   & \textbf{9.18$^\ast$}    & \textbf{1.50$^\ast$}   & \textbf{1.00$^\ast$}  & \textbf{21.70$^\ast$}   & \textbf{11.96$^\ast$}   & \textbf{9.22$^\ast$}  \\
      &     & TCL  & 9.38   & 1.31   & 0.30   & 20.40  & 9.74   & 6.80  \\
      &     & CLIP$_{\rm ViT}$    & 20.98   & 7.79   & 4.57   & 31.73   & 19.13   & 14.63  \\
      &     & CLIP$_{\rm CNN}$    & 22.48   & 7.40  & 4.12  & 31.94    & 21.36  & 15.59  \\
      \cline{2-9}
      & \multirow{4}{*}{\texttt{\textbf{Image}@Multi}}       & ALBEF   & \textbf{75.50$^\ast$}    & \textbf{59.22$^\ast$}   & \textbf{53.30$^\ast$}   & \textbf{83.63$^\ast$}   & \textbf{75.14$^\ast$}   & \textbf{70.32$^\ast$}  \\
      &     & TCL  & 4.64   & 1.21   & 0.50   & 11.33  & 3.72   & 2.25  \\
      &     & CLIP$_{\rm ViT}$    & 7.24   & 1.97   & 0.51   & 13.53   & 5.23   & 3.01  \\
      &     & CLIP$_{\rm CNN}$    & 10.09   & 2.85  & 1.65  & 15.27    & 6.11  & 3.52  \\
      \cline{2-9}
      & \multirow{4}{*}{\texttt{\textbf{Bi}@Multi}}       & ALBEF   & \textbf{77.16$^\ast$}    & \textbf{64.60$^\ast$}   & \textbf{58.37$^\ast$}   & \textbf{83.86$^\ast$}   & \textbf{74.63$^\ast$}   & \textbf{70.13$^\ast$}  \\
      &     & TCL  & 15.21   & 4.19   & 1.47   & 29.49  & 14.97   & 10.55  \\
      &     & CLIP$_{\rm ViT}$    & 23.60   & 7.82   & 3.93   & 36.48   & 21.09   & 15.76  \\
      &     & CLIP$_{\rm CNN}$    & 25.12   & 8.42  & 5.39  & 38.89    & 22.38  & 17.49  \\
        \hline
    \multirow{12}{*}{\textbf{TCL}}                      & \multirow{4}{*}{\texttt{\textbf{Text}@Multi}}       
    & TCL   & \textbf{12.86}$^\ast$ &\textbf{2.81}$^\ast$ &\textbf{1.0}$^\ast$ &\textbf{30.33}$^\ast$ &\textbf{15.32}$^\ast$ &\textbf{10.89}$^\ast$  \\
      &     & ALBEF  & 13.24 & 2.61 & 1.2 & 27.13 & 15.16 & 11.28  \\
      &     & CLIP$_{\rm ViT}$    & 25.28 & 9.87 & 5.79 & 37.11 & 22.85 & 17.25  \\
      &     & CLIP$_{\rm CNN}$    & 26.18 & 11.21 & 5.25 & 37.84 & 24.65 & 18.71 \\
      \cline{2-9}
      & \multirow{4}{*}{\texttt{\textbf{Image}@Multi}}       
      & TCL   & \textbf{72.5}$^\ast$ &\textbf{55.98}$^\ast$ &\textbf{46.49}$^\ast$ &\textbf{79.26}$^\ast$ &\textbf{64.65}$^\ast$ &\textbf{56.99}$^\ast$  \\
      &     & ALBEF  & 5.94 & 1.6 & 0.8 & 12.16 & 3.79 & 2.26  \\
      &     & CLIP$_{\rm ViT}$    & 7.85 & 1.97 & 0.61 & 13.43 & 5.37 & 3.31  \\
      &     & CLIP$_{\rm CNN}$    & 9.71 & 2.85 & 1.65 & 15.44 & 5.7 & 3.37  \\
      \cline{2-9}
      & \multirow{4}{*}{\texttt{\textbf{Bi}@Multi}}       
      & TCL   & \textbf{78.08}$^\ast$ &\textbf{65.53}$^\ast$ &\textbf{56.81}$^\ast$ &\textbf{87.43}$^\ast$ &\textbf{75.23}$^\ast$ &\textbf{68.87}$^\ast$  \\
      &     & ALBEF  & 22.94 & 6.61 & 3.6 & 40.13 & 22.72 & 17.51  \\
      &     & CLIP$_{\rm ViT}$    & 27.98 & 9.66 & 5.08 & 41.46 & 25.11 & 18.99  \\
      &     & CLIP$_{\rm CNN}$    & 30.78 & 12.47 & 7.52 & 44.19 & 26.93 & 20.63  \\
        \hline
      \multirow{12}{*}{\textbf{CLIP$_{\rm ViT}$}}                      & \multirow{4}{*}{\texttt{\textbf{Text}@Uni}}       & CLIP$_{\rm ViT}$   & \textbf{28.34$^\ast$}    & \textbf{11.73$^\ast$}   & \textbf{6.81$^\ast$}   & \textbf{38.89$^\ast$}   & \textbf{24.08$^\ast$}   & \textbf{17.42$^\ast$}  \\
      &     & CLIP$_{\rm CNN}$  & 29.89   & 11.52   & 5.87   & 37.36  & 24.97  & 18.62  \\
      &     & ALBEF    & 7.61   & 1.00   & 0.30   & 19.97   & 9.58   & 6.59  \\
      &     &  TCL   & 8.43   & 0.90  & 0.30  & 20.90    & 9.96  & 7.03  \\
      \cline{2-9}
      & \multirow{4}{*}{\texttt{\textbf{Image}@Uni}}       & CLIP$_{\rm ViT}$   & \textbf{87.73$^\ast$}    & \textbf{78.09$^\ast$}   & \textbf{72.05$^\ast$}   & \textbf{91.72$^\ast$}   & \textbf{83.32$^\ast$}   & \textbf{78.67$^\ast$}  \\
      &     & CLIP$_{\rm CNN}$  & 7.66  & 1.90   & 1.44   & 9.37  & 3.90   & 2.53  \\
      &     & ALBEF    & 2.50   & 0.60   & 0.20   & 5.80   & 1.78   & 1.11  \\
      &     &  TCL   & 5.27  & 0.40  & 0.20  & 9.12   & 2.75  & 1.48  \\
      \cline{2-9}
      & \multirow{4}{*}{\texttt{\textbf{Bi}@Uni}}       & CLIP$_{\rm ViT}$   & \textbf{93.25$^\ast$}    & \textbf{84.88$^\ast$}   & \textbf{78.96$^\ast$}   & \textbf{95.86$^\ast$}   & \textbf{90.83$^\ast$}   & \textbf{87.36$^\ast$}  \\
      &     & CLIP$_{\rm CNN}$  & 32.52   & 13.78   & 7.52   & 41.82  & 26.77   & 21.10  \\
      &     & ALBEF    & 10.57   & 1.87   & 0.63   & 24.33   & 11.74   & 8.41  \\
      &     &  TCL   & 11.94   & 2.38  & 1.07  & 26.69    & 13.80  & 9.46  \\
        \hline
    \multirow{12}{*}{\textbf{CLIP$_{\rm CNN}$}}                      & \multirow{4}{*}{\texttt{\textbf{Text}@Uni}}       & CLIP$_{\rm CNN}$   & \textbf{30.40$^\ast$}    & \textbf{13.11$^\ast$}   & \textbf{7.21$^\ast$}   & \textbf{40.03$^\ast$}   & \textbf{26.79$^\ast$}   & \textbf{20.74$^\ast$}  \\
      &     & CLIP$_{\rm ViT}$  & 26.99   & 11.11   & 6.81   & 37.37  & 23.48  & 17.64  \\
      &     & ALBEF    & 7.72   & 0.90   & 0.50   & 20.79   & 9.84   & 6.98  \\
      &     &  TCL   & 9.69   & 1.31  & 0.30  & 21.67    & 10.73  & 7.49  \\
      \cline{2-9}
      & \multirow{4}{*}{\texttt{\textbf{Image}@Uni}}       & CLIP$_{\rm CNN}$   & \textbf{88.12$^\ast$}    & \textbf{79.70$^\ast$}   & \textbf{74.87$^\ast$}   & \textbf{93.69$^\ast$}   & \textbf{87.66$^\ast$}   & \textbf{83.03$^\ast$}  \\
      &     & CLIP$_{\rm ViT}$  & 1.84  & 0.10   & 0.30   & 5.51  & 2.50   & 1.02  \\
      &     & ALBEF    & 1.98   & 0.30   & 0.20   & 5.12  & 1.42   & 0.91  \\
      &     &  TCL   & 4.74   & 0.50  & 0.10  & 7.95    & 2.32  & 1.42  \\
      \cline{2-9}
      & \multirow{4}{*}{\texttt{\textbf{Bi}@Uni}}       & CLIP$_{\rm CNN}$   & \textbf{94.76$^\ast$}    & \textbf{87.03$^\ast$}   & \textbf{82.08$^\ast$}   & \textbf{96.89$^\ast$}   & \textbf{92.87$^\ast$}   & \textbf{89.25$^\ast$}  \\
      &     & CLIP$_{\rm ViT}$  & 28.79   & 11.63   & 6.40   & 40.03  & 24.60  & 18.83  \\
      &     & ALBEF    & 8.79  & 1.53   & 0.60   & 23.74   & 11.75   & 8.42  \\
      &     &  TCL   & 13.10   & 2.31  & 0.93  & 26.07    & 13.53  & 9.23  \\
        \bottomrule[0.3mm]
    \end{tabular}}
\end{center}
\caption{
\textbf{Attack success rates ($\%$)} with different adversarial input modalities under Co-Attack on image-text retrieval. The adversaries are crafted using Flickr30K. $^\ast$ indicates white-box attacks. A higher ASR indicates better adversarial transferability.}
\vspace{-5pt}
\label{tab:supp_t11_exp_co_attack}
\end{table*}

\begin{sidewaystable*}[t]
	\centering
	\footnotesize 
	\renewcommand\arraystretch{0.9}
\setlength{\tabcolsep}{2pt}
		\scalebox{0.94}[0.94]{
		\begin{tabular}{ @{\extracolsep{\fill}} l|l|ccc|ccc|ccc|ccc} 
        \toprule[0.3mm]
        \multicolumn{14}{c}{\textbf{Flickr30K  (Image-Text Retrieval)}} \\ \midrule[0.3mm]
			& &  \multicolumn{3}{c}{\textbf{ALBEF}} & \multicolumn{3}{c}{\textbf{TCL}} & \multicolumn{3}{c}{\textbf{CLIP$_{\rm ViT}$}} & \multicolumn{3}{c}{\textbf{CLIP$_{\rm CNN}$}}  \\
			\cmidrule{3-14}
			\multirow{-2}{*}{\textbf{Source}} &\multirow{-2}{*}{\textbf{Attack}} & {R@1} & {R@5} & {R@10} & {R@1} & {R@5} & {R@10} & {R@1} & {R@5} & {R@10} & {R@1} & {R@5} & {R@10} \\
			\midrule
			\multirow{5}{*}{\rotatebox[origin=c]{0}{\textbf{ALBEF}}} 
            & PGD 
                & 52.45$^\ast$ & 36.57$^\ast$ & 30.00$^\ast$   
                & 3.06 & 0.40 & 0.10   
                & 8.96 & 1.66 & 0.41   
                & 10.34 & 2.96 & 1.85  \\
            & BERT-Attack 
                & 11.57$^\ast$ & 1.80$^\ast$ & 1.10$^\ast$  
                & 12.64 & 2.51 & 0.90 
                & 29.33 & 11.63 & 6.30 
                & 32.69 & 15.43 & 8.65  \\
            & Sep-Attack 
                & 65.69$^\ast$ & 47.60$^\ast$ & 42.10$^\ast$
                & 17.60 & 3.72 & 1.90 
                & 31.17 & 12.05 & 7.01
                & 32.82 & 15.86 & 9.06  \\
            & Co-Attack 
                & 77.16$^\ast$  & 64.60$^\ast$  & 58.37$^\ast$  
                & 15.21 & 4.19 & 1.47 
                & 23.60 & 7.82 & 3.93 
                & 25.12 & 8.42 & 5.39 \\
			& \cellcolor{gray! 20} SGA  
                & \cellcolor{gray! 20}\textbf{97.24$\pm$0.22$^\ast$} 
                & \cellcolor{gray! 20}\textbf{94.09$\pm$0.42$^\ast$} 
                & \cellcolor{gray! 20}\textbf{92.30$\pm$0.28$^\ast$}  
                & \cellcolor{gray! 20}\textbf{45.42$\pm$0.60} 
                & \cellcolor{gray! 20}\textbf{24.93$\pm$0.15} 
                & \cellcolor{gray! 20}\textbf{16.48$\pm$0.49} 
                & \cellcolor{gray! 20}\textbf{33.38$\pm$0.35} 
                & \cellcolor{gray! 20}\textbf{13.50$\pm$0.30} 
                & \cellcolor{gray! 20}\textbf{9.04$\pm$0.15} 
                & \cellcolor{gray! 20}\textbf{34.93$\pm$0.99} 
                & \cellcolor{gray! 20}\textbf{17.07$\pm$0.23} 
                & \cellcolor{gray! 20}\textbf{10.45$\pm$0.95} \\
			\midrule
			\multirow{5}{*}{\rotatebox[origin=c]{0}{\textbf{TCL}}} 
            & PGD 
                & 6.15 & 1.30 & 0.70
                & 77.87$^\ast$ & 65.13$^\ast$ & 58.72$^\ast$
                & 7.48 & 1.45 & 0.81
                & 10.34 & 2.75 & 1.54 \\
            & BERT-Attack 
                & 11.89 & 2.20 & 0.70
                & 14.54$^\ast$ & 2.31$^\ast$ & 0.60$^\ast$
                & 29.69 & 12.77 & 7.62
                & 33.46 & 14.38 & 9.37 \\
            & Sep-Attack    
                & 20.13 & 4.91 & 2.70 
                & 84.72$^\ast$ & 73.07$^\ast$ & 65.43$^\ast$
                & 31.29 & 12.98 & 7.72
                & 33.33 & 14.27 & 9.89 \\
            & Co-Attack 
                & 23.15 & 6.98  & 3.63
                & 77.94$^\ast$  & 64.26$^\ast$  & 56.18$^\ast$  
                & 27.85 & 9.80  & 5.22 
                & 30.74 & 12.09 & 7.28 \\
			& \cellcolor{gray! 20} SGA  
                & \cellcolor{gray! 20}\textbf{48.91$\pm$0.74} 
                & \cellcolor{gray! 20}\textbf{30.86$\pm$0.28} 
                & \cellcolor{gray! 20}\textbf{23.10$\pm$0.42} 
                & \cellcolor{gray! 20}\textbf{98.37$\pm$0.08$^\ast$}  
                & \cellcolor{gray! 20}\textbf{96.53$\pm$0.07$^\ast$} 
                & \cellcolor{gray! 20}\textbf{94.99$\pm$0.28$^\ast$} 
                & \cellcolor{gray! 20}\textbf{33.87$\pm$0.18} 
                & \cellcolor{gray! 20}\textbf{15.21$\pm$0.07} 
                & \cellcolor{gray! 20}\textbf{9.46$\pm$0.43} 
                & \cellcolor{gray! 20}\textbf{37.74$\pm$0.27} 
                & \cellcolor{gray! 20}\textbf{17.86$\pm$0.30} 
                & \cellcolor{gray! 20}\textbf{11.74$\pm$0.00}  \\
			\midrule
			\multirow{5}{*}{\rotatebox[origin=c]{0}{\textbf{CLIP$_{\rm ViT}$}}} 

            & PGD 
                & 2.50 & 0.40 & 0.10
                & 4.85 & 0.20 & 0.20
                & 70.92$^\ast$ & 50.05$^\ast$ & 42.28$^\ast$
                & 5.36 & 1.16 & 0.72 \\
            & BERT-Attack 
                & 9.59 & 1.30 & 0.40
                & 11.80 & 1.91 & 0.70
                & 28.34$^\ast$ & 11.73$^\ast$ & 6.81$^\ast$
                & 30.40 & 11.63 & 5.97 \\
            & Sep-Attack        
                & 9.59 & 1.30 & 0.50 & 11.38 & 2.11 & 0.90 & 79.75$^\ast$ & 63.03$^\ast$ & 53.76$^\ast$ & 30.78 & 12.16 & 6.39 \\ 
            & Co-Attack 
                & 10.57 & 1.87 & 0.63
                & 11.94 & 2.38 & 1.07
                & 93.25$^\ast$ & 84.88$^\ast$ & 78.96$^\ast$ 
                & 32.52 & 13.78 & 7.52  \\
			& \cellcolor{gray! 20}SGA  
                & \cellcolor{gray! 20}\textbf{13.40$\pm$0.07} 
                & \cellcolor{gray! 20}\textbf{2.46$\pm$0.08} 
                & \cellcolor{gray! 20}\textbf{1.35$\pm$0.07} 
                & \cellcolor{gray! 20}\textbf{16.23$\pm$0.45} 
                & \cellcolor{gray! 20}\textbf{3.77$\pm$0.21} 
                & \cellcolor{gray! 20}\textbf{1.10$\pm$0.14} 
                & \cellcolor{gray! 20}\textbf{99.08$\pm$0.08$^\ast$} 
                & \cellcolor{gray! 20}\textbf{97.25$\pm$0.07$^\ast$} 
                & \cellcolor{gray! 20}\textbf{95.22$\pm$0.15$^\ast$} 
                & \cellcolor{gray! 20}\textbf{38.76$\pm$0.27}
                & \cellcolor{gray! 20}\textbf{19.45$\pm$0.00} 
                & \cellcolor{gray! 20}\textbf{11.95$\pm$0.44}  \\
			\midrule
			\multirow{5}{*}{\rotatebox[origin=c]{0}{\textbf{CLIP$_{\rm CNN}$}}} 
            & PGD 
                & 2.09 & 0.30 & 0.10 & 4.00 & 0.40 & 0.20 & 1.10 & 0.52 & 0.41 & 86.46$^\ast$ & 69.13$^\ast$ & 61.17$^\ast$ \\
            & BERT-Attack 
                & 8.86 & 1.50 & 0.60 & 12.33 & 2.01 & 0.90 & 27.12 & 11.21 & 6.81 & 30.40$^\ast$ & 13.00$^\ast$ & 7.31$^\ast$ \\
            & Sep-Attack        
                & 8.55 & 1.50 & 0.60 & 12.64 & 1.91 & 0.70 & 28.34 & 10.8 & 6.30 & 91.44$^\ast$ & 78.54$^\ast$ & 71.58$^\ast$ \\
            & Co-Attack 
                & 8.79 & 1.53 & 0.60 
                & 13.10 & 2.31 & 0.93 
                & 28.79 & 11.63 & 6.40 
                & 94.76$^\ast$ & 87.03$^\ast$ & 82.08$^\ast$ \\
			& \cellcolor{gray! 20}SGA  
                & \cellcolor{gray! 20}\textbf{11.42$\pm$0.07} 
                & \cellcolor{gray! 20}\textbf{2.56$\pm$0.07} 
                & \cellcolor{gray! 20}\textbf{1.05$\pm$0.21} 
                & \cellcolor{gray! 20}\textbf{14.91$\pm$0.08} 
                & \cellcolor{gray! 20}\textbf{3.62$\pm$0.14} 
                & \cellcolor{gray! 20}\textbf{1.70$\pm$0.14} 
                & \cellcolor{gray! 20}\textbf{31.24$\pm$0.42} 
                & \cellcolor{gray! 20}\textbf{13.45$\pm$0.07} 
                & \cellcolor{gray! 20}\textbf{8.74$\pm$0.14} 
                & \cellcolor{gray! 20}\textbf{99.24$\pm$0.18$^\ast$} 
                & \cellcolor{gray! 20}\textbf{98.20$\pm$0.30$^\ast$} 
                & \cellcolor{gray! 20}\textbf{95.16$\pm$0.44$^\ast$}   \\
   \midrule[0.3mm]
			\multicolumn{14}{c}{\textbf{Flickr30K (Text-Image Retrieval)}} \\ \midrule[0.3mm]
			\multirow{5}{*}{\rotatebox[origin=c]{0}{\textbf{ALBEF}}} 
            & PGD 
                & 58.65$^\ast$ & 44.85$^\ast$ & 38.98$^\ast$   
                & 6.79 & 2.21 & 1.20   
                & 13.21 & 5.19 & 3.05   
                & 14.65 & 5.60 & 3.39 \\
            & BERT-Attack 
                & 27.46$^\ast$ & 14.48$^\ast$ & 10.98$^\ast$
                & 28.07 & 14.39 & 10.26
                & 43.17 & 26.37 & 19.91
                & 46.11 & 28.43 & 22.14 \\
            & Sep-Attack 
                & 73.95$^\ast$ & 59.50$^\ast$ & 53.70$^\ast$
                & 32.95 & 17.10 & 11.90
                & \textbf{45.23} & 25.93 & 19.95
                & 45.49 & 28.43 & 22.32 \\
            & Co-Attack 
                & 83.86$^\ast$ & 74.63$^\ast$ & 70.13$^\ast$ 
                & 29.49 & 14.97 & 10.55 
                & 36.48 & 21.09 & 15.76 
                & 38.89 & 22.38 & 17.49  \\
			& \cellcolor{gray! 20}SGA  
                & \cellcolor{gray! 20}\textbf{97.28$\pm$0.15$^\ast$} 
                & \cellcolor{gray! 20}\textbf{94.27$\pm$0.04$^\ast$} 
                & \cellcolor{gray! 20}\textbf{92.58$\pm$0.03$^\ast$} 
                & \cellcolor{gray! 20}\textbf{55.25$\pm$0.06} 
                & \cellcolor{gray! 20}\textbf{36.01$\pm$0.03} 
                & \cellcolor{gray! 20}\textbf{27.25$\pm$0.13} 
                & \cellcolor{gray! 20} 44.16$\pm$0.25 
                & \cellcolor{gray! 20}\textbf{27.35$\pm$0.30} 
                & \cellcolor{gray! 20}\textbf{20.84$\pm$0.04} 
                & \cellcolor{gray! 20}\textbf{46.57$\pm$0.13} 
                & \cellcolor{gray! 20}\textbf{29.16$\pm$0.17} 
                & \cellcolor{gray! 20}\textbf{22.68$\pm$0.00} \\
			\midrule
			\multirow{5}{*}{\rotatebox[origin=c]{0}{\textbf{TCL}}} 
            & PGD 
                & 10.78 & 3.36 & 1.70
                & 79.48$^\ast$ & 66.26$^\ast$ & 60.36$^\ast$
                & 13.72 & 5.37 & 3.01
                & 15.33 & 5.77 & 3.28 \\
            & BERT-Attack 
                & 26.82 & 14.09 & 10.80
                & 29.17$^\ast$ & 15.03$^\ast$ & 10.91$^\ast$
                & 44.49 & 27.47 & 21.00
                & 46.07 & 29.28 & 22.59 \\
            & Sep-Attack  
                & 36.48 & 19.48 & 14.82
                & 86.07$^\ast$ & 74.67$^\ast$ & 68.83$^\ast$
                & 44.65 & 26.82 & 20.37
                & 45.80 & 29.18 & 23.02 \\
            & Co-Attack 
                & 40.04 & 22.66 & 17.23 & 85.59$^\ast$ & 74.19$^\ast$ & 68.25$^\ast$ & 41.19 & 25.22 & 19.01 & 44.11 & 26.67 & 20.66 \\
			& \cellcolor{gray! 20}SGA 
                & \cellcolor{gray! 20}\textbf{60.34$\pm$0.10} 
                & \cellcolor{gray! 20}\textbf{42.47$\pm$0.22} 
                & \cellcolor{gray! 20}\textbf{34.59$\pm$0.29} 
                & \cellcolor{gray! 20}\textbf{98.81$\pm$0.07$^\ast$} 
                & \cellcolor{gray! 20}\textbf{97.19$\pm$0.03$^\ast$} 
                & \cellcolor{gray! 20}\textbf{95.86$\pm$0.11$^\ast$} 
                & \cellcolor{gray! 20}\textbf{44.88$\pm$0.54} 
                & \cellcolor{gray! 20}\textbf{28.79$\pm$0.28} 
                & \cellcolor{gray! 20}\textbf{21.95$\pm$0.11} 
                & \cellcolor{gray! 20}\textbf{48.30$\pm$0.34} 
                & \cellcolor{gray! 20}\textbf{29.70$\pm$0.02} 
                & \cellcolor{gray! 20}\textbf{23.68$\pm$0.06}   \\
			\midrule
			\multirow{5}{*}{\rotatebox[origin=c]{0}{\textbf{CLIP$_{\rm ViT}$}}} 
            & PGD 
                & 4.93 & 1.44 & 1.01
                & 8.17 & 2.27 & 1.46
                & 78.61$^\ast$ & 60.78$^\ast$ & 51.50$^\ast$
                & 8.44 & 2.35 & 1.54 \\
            & BERT-Attack 
                & 22.64 & 10.95 & 8.17
                & 25.07 & 12.92 & 8.90
                & 39.08$^\ast$ & 24.08$^\ast$ & 17.44$^\ast$
                & 37.43 & 24.96 & 18.66  \\
            & Sep-Attack      
                & 23.25 & 11.22 & 8.01 & 25.60 & 12.92 & 9.14 & 86.79$^\ast$ & 75.24$^\ast$ & 67.84$^\ast$ & 39.76 & 25.62 & 19.34 \\
            & Co-Attack 
                & 24.33 & 11.74 & 8.41 
                & 26.69 & 13.80 & 9.46
                & 95.86$^\ast$ & 90.83$^\ast$ & 87.36$^\ast$ 
                & 41.82 & 26.77 & 21.10 \\
			& \cellcolor{gray! 20}SGA  
                & \cellcolor{gray! 20}\textbf{27.22$\pm$0.06} 
                & \cellcolor{gray! 20}\textbf{13.21$\pm$0.00} 
                & \cellcolor{gray! 20}\textbf{9.76$\pm$0.11} 
                & \cellcolor{gray! 20}\textbf{30.76$\pm$0.07} 
                & \cellcolor{gray! 20}\textbf{16.36$\pm$0.26} 
                & \cellcolor{gray! 20}\textbf{12.08$\pm$0.06} 
                & \cellcolor{gray! 20}\textbf{98.94$\pm$0.00$^\ast$} 
                & \cellcolor{gray! 20}\textbf{97.53$\pm$0.16$^\ast$} 
                & \cellcolor{gray! 20}\textbf{96.03$\pm$0.08$^\ast$} 
                & \cellcolor{gray! 20}\textbf{47.79$\pm$0.58} 
                & \cellcolor{gray! 20}\textbf{30.36$\pm$0.36} 
                & \cellcolor{gray! 20}\textbf{24.50$\pm$0.37}  \\
			\midrule
			\multirow{5}{*}{\rotatebox[origin=c]{0}{\textbf{CLIP$_{\rm CNN}$}}} 
             & PGD 
                & 4.82 & 1.29 & 0.87 & 7.81 & 2.09 & 1.34 & 6.60 & 2.73 & 1.48 & 92.25$^\ast$ & 81.00$^\ast$ & 75.04$^\ast$ \\
            & BERT-Attack 
                & 23.27 & 11.34 & 8.41 & 25.48 & 13.25 & 8.81 & 37.44 & 23.48 & 17.66 & 40.10$^\ast$ & 26.71$^\ast$ & 20.85$^\ast$ \\
            & Sep-Attack        
                & 23.41 & 11.38 & 8.23 & 26.12 & 13.44 & 8.96 & 39.43 & 24.34 & 18.36 & 95.44$^\ast$ & 88.48$^\ast$ & 82.88$^\ast$ \\
            & Co-Attack & 23.74 & 11.75 & 8.42 & 26.07 & 13.53 & 9.23 & 40.03 & 24.60 & 18.83 & 96.89$^\ast$ & 92.87$^\ast$ & 89.25$^\ast$ \\
			& \cellcolor{gray! 20}SGA  
                & \cellcolor{gray! 20}\textbf{24.80$\pm$0.28} 
                & \cellcolor{gray! 20}\textbf{12.32$\pm$0.15} 
                & \cellcolor{gray! 20}\textbf{8.98$\pm$0.06}
                & \cellcolor{gray! 20}\textbf{28.82$\pm$0.11} 
                & \cellcolor{gray! 20}\textbf{15.12$\pm$0.11} 
                & \cellcolor{gray! 20}\textbf{10.56$\pm$0.17} 
                & \cellcolor{gray! 20}\textbf{42.12$\pm$0.11} 
                & \cellcolor{gray! 20}\textbf{26.80$\pm$0.05} 
                & \cellcolor{gray! 20}\textbf{20.23$\pm$0.13} 
                & \cellcolor{gray! 20}\textbf{99.49$\pm$0.05$^\ast$} 
                & \cellcolor{gray! 20}\textbf{98.41$\pm$0.06$^\ast$} 
                & \cellcolor{gray! 20}\textbf{97.14$\pm$0.11$^\ast$}  \\
			\bottomrule[0.3mm]		
	\end{tabular}}
	\caption{\textbf{Attack success rate ($\%$)} of four VLP models under existing adversarial attacks and SGA.
	The source column indicates the source models used to generate the adversarial data on Flickr30K.
	 $^\ast$ indicates white-box attacks. A higher ASR indicates better adversarial transferability.}
	\vspace{-8pt}
	\label{tab:supp_t12_SGA_flickr}
\end{sidewaystable*}
\begin{sidewaystable*}[t]
	\centering
	\footnotesize
	\renewcommand\arraystretch{0.9}
\setlength{\tabcolsep}{2pt}
		\scalebox{0.94}[0.94]{
		\begin{tabular}{ @{\extracolsep{\fill}} l|l|ccc|ccc|ccc|ccc}
        \toprule[0.3mm]
        \multicolumn{14}{c}{\textbf{MSCOCO  (Image-Text Retrieval)}} \\ \midrule[0.3mm]
			& &  \multicolumn{3}{c}{\textbf{ALBEF}} & \multicolumn{3}{c}{\textbf{TCL}} & \multicolumn{3}{c}{\textbf{CLIP$_{\rm ViT}$}} & \multicolumn{3}{c}{\textbf{CLIP$_{\rm CNN}$}}  \\
			\cmidrule{3-14}
			\multirow{-2}{*}{\textbf{Source}} &\multirow{-2}{*}{\textbf{Attack}} & {R@1} & {R@5} & {R@10} & {R@1} & {R@5} & {R@10} & {R@1} & {R@5} & {R@10} & {R@1} & {R@5} & {R@10} \\
			\midrule
			\multirow{5}{*}{\rotatebox[origin=c]{0}{\textbf{ALBEF}}} 
            & PGD & 76.70$^\ast$ & 67.49$^\ast$ & 62.47$^\ast$ & 12.46 & 5.00 & 3.14 & 13.96 & 7.33 & 5.21 & 17.45 & 9.08 & 6.45 \\
& BERT-Attack & 24.39$^\ast$ & 10.67$^\ast$ & 6.75$^\ast$ & 24.34 & 9.92 & 6.25 & 44.94 & 27.97 & 22.55 & 47.73 & 29.56 & 23.10 \\
& Sep-Attack & 82.60$^\ast$ & 73.20$^\ast$ & 67.58$^\ast$ & 32.83 & 15.52 & 10.10 & 44.03 & 27.60 & 21.84 & 46.96 & 29.83 & 23.15 \\
            & Co-Attack  & 79.87$^\ast$ & 68.62$^\ast$ & 62.88$^\ast$ & 32.62 & 15.36 & 9.67 & 44.89 & 28.33 & 21.89 & 47.30 & 29.89 & 23.29  \\
			& \cellcolor{gray! 20}SGA  & \cellcolor{gray! 20}\textbf{96.75$\pm$0.11$^\ast$} & \cellcolor{gray! 20}\textbf{92.83$\pm$0.13$^\ast$} & \cellcolor{gray! 20}\textbf{90.37$\pm$0.03$^\ast$} & \cellcolor{gray! 20}\textbf{58.56$\pm$0.06} & \cellcolor{gray! 20}\textbf{39.00$\pm$0.40} & \cellcolor{gray! 20}\textbf{30.68$\pm$0.22} & \cellcolor{gray! 20}\textbf{57.06$\pm$0.51} & \cellcolor{gray! 20}\textbf{39.38$\pm$0.22} & \cellcolor{gray! 20}\textbf{31.55$\pm$0.06} & \cellcolor{gray! 20}\textbf{58.95$\pm$0.19} & \cellcolor{gray! 20}\textbf{42.49$\pm$0.13} & \cellcolor{gray! 20}\textbf{34.84$\pm$0.28}  \\
			\midrule
			\multirow{5}{*}{\rotatebox[origin=c]{0}{\textbf{TCL}}} 
            & PGD & 10.83 & 5.28 & 3.21 & 59.58$^\ast$ & 51.25$^\ast$ & 47.89$^\ast$ & 14.23 & 7.40 & 4.93 & 17.25 & 8.51 & 6.45 \\
            & BERT-Attack & 35.32 & 15.89 & 10.25 & 38.54$^\ast$ & 19.08$^\ast$ & 12.10$^\ast$ & 51.09 & 31.71 & 25.40 & 52.23 & 33.75 & 27.06 \\
            & Sep-Attack & 41.71 & 21.37 & 14.99 & 70.32$^\ast$ & 59.64$^\ast$ & 55.09$^\ast$ & 50.74 & 31.34 & 24.43 & 51.90 & 34.02 & 26.79 \\
            &  Co-Attack  & 46.08 & 24.87 & 17.11 & 85.38$^\ast$ & 74.73$^\ast$ & 68.23$^\ast$ & 51.62 & 31.92 & 24.87 & 52.13 & 33.80 & 27.09 \\
			& \cellcolor{gray! 20}SGA  & \cellcolor{gray! 20}\textbf{65.93$\pm$0.06} & \cellcolor{gray! 20}\textbf{49.33$\pm$0.35} & \cellcolor{gray! 20}\textbf{40.34$\pm$0.01} & \cellcolor{gray! 20}\textbf{98.97$\pm$0.04$^\ast$} & \cellcolor{gray! 20}\textbf{97.89$\pm$0.12$^\ast$} & \cellcolor{gray! 20}\textbf{96.63$\pm$0.03$^\ast$} & \cellcolor{gray! 20}\textbf{56.34$\pm$0.08} & \cellcolor{gray! 20}\textbf{39.58$\pm$0.21} & \cellcolor{gray! 20}\textbf{32.00$\pm$0.12} & \cellcolor{gray! 20}\textbf{59.44$\pm$0.20} & \cellcolor{gray! 20}\textbf{42.17$\pm$0.21} & \cellcolor{gray! 20}\textbf{34.94$\pm$0.05}  \\
			\midrule
			\multirow{5}{*}{\rotatebox[origin=c]{0}{\textbf{CLIP$_{\rm ViT}$}}} 
& PGD & 7.24 & 3.10 & 1.65 & 10.19 & 4.23 & 2.50 & 54.79$^\ast$ & 36.21$^\ast$ & 28.57$^\ast$ & 7.32 & 3.64 & 2.79 \\
& BERT-Attack & 20.34 & 8.53 & 4.73 & 21.08 & 7.96 & 4.65 & 45.06$^\ast$ & 28.62$^\ast$ & 22.67$^\ast$ & 44.54 & 29.37 & 23.97 \\
& Sep-Attack & 23.41 & 10.33 & 6.15 & 25.77 & 11.60 & 7.45 & 68.52$^\ast$ & 52.30$^\ast$ & 43.88$^\ast$ & 43.11 & 27.22 & 21.77 \\
             &  Co-Attack  & 30.28 & 13.64 & 8.83 & 32.84 & 15.27 & 10.27 & 97.98$^\ast$ & 94.94$^\ast$ & 93.00$^\ast$ & 55.08 & 38.64 & 31.42  \\
			& \cellcolor{gray! 20}SGA  & \cellcolor{gray! 20}\textbf{33.41$\pm$0.22} & \cellcolor{gray! 20}\textbf{16.73$\pm$0.04} & \cellcolor{gray! 20}\textbf{10.98$\pm$0.25} & \cellcolor{gray! 20}\textbf{37.54$\pm$0.30} & \cellcolor{gray! 20}\textbf{19.09$\pm$0.04} & \cellcolor{gray! 20}\textbf{12.92$\pm$0.31} & \cellcolor{gray! 20}\textbf{99.79$\pm$0.03$^\ast$} & \cellcolor{gray! 20}\textbf{99.37$\pm$0.07$^\ast$} & \cellcolor{gray! 20}\textbf{98.89$\pm$0.04$^\ast$} & \cellcolor{gray! 20}\textbf{58.93$\pm$0.11} & \cellcolor{gray! 20}\textbf{44.60$\pm$0.11} & \cellcolor{gray! 20}\textbf{37.53$\pm$0.74}  \\
			\midrule
			\multirow{5}{*}{\rotatebox[origin=c]{0}{\textbf{CLIP$_{\rm CNN}$}}} 
            & PGD & 7.01 & 3.03 & 1.77 & 10.08 & 4.20 & 2.38 & 4.88 & 2.96 & 1.71 & 76.99$^\ast$ & 63.80$^\ast$ & 56.76$^\ast$ \\
& BERT-Attack & 23.38 & 10.16 & 5.70 & 24.58 & 9.70 & 5.96 & 51.28 & 33.23 & 26.63 & 54.43$^\ast$ & 38.26$^\ast$ & 30.74$^\ast$ \\
& Sep-Attack & 26.53 & 11.78 & 6.88 & 30.26 & 13.00 & 8.61 & 50.44 & 32.71 & 25.92 & 88.72$^\ast$ & 78.71$^\ast$ & 72.77$^\ast$ \\
            &  Co-Attack  & 29.83 & 13.13 & 8.35 & 32.97 & 15.11 & 9.76 & 53.10 & 35.91 & 28.53 & 96.72$^\ast$ & 94.02$^\ast$ & 91.57$^\ast$ \\
			& \cellcolor{gray! 20}SGA  & \cellcolor{gray! 20}\textbf{31.61$\pm$0.40} & \cellcolor{gray! 20}\textbf{14.27$\pm$0.28} & \cellcolor{gray! 20}\textbf{9.36$\pm$0.01} & \cellcolor{gray! 20}\textbf{34.81$\pm$0.15} & \cellcolor{gray! 20}\textbf{17.16$\pm$0.03} & \cellcolor{gray! 20}\textbf{11.26$\pm$0.04} & \cellcolor{gray! 20}\textbf{56.62$\pm$0.06} & \cellcolor{gray! 20}\textbf{41.31$\pm$0.15} & \cellcolor{gray! 20}\textbf{32.88$\pm$0.10} & \cellcolor{gray! 20}\textbf{99.61$\pm$0.08$^\ast$} & \cellcolor{gray! 20}\textbf{99.02$\pm$0.11$^\ast$} & \cellcolor{gray! 20}\textbf{98.42$\pm$0.17$^\ast$}  \\
			\hline \midrule[0.3mm]
			\multicolumn{14}{c}{\textbf{MSCOCO (Text-Image Retrieval)}} \\ \midrule[0.3mm]
			\multirow{5}{*}{\rotatebox[origin=c]{0}{\textbf{ALBEF}}} 
            & PGD & 86.30$^\ast$ & 78.49$^\ast$ & 73.94$^\ast$ & 17.77 & 8.36 & 5.32 & 23.10 & 12.74 & 9.43 & 23.54 & 13.26 & 9.61 \\
& BERT-Attack & 36.13$^\ast$ & 23.71$^\ast$ & 18.94$^\ast$ & 33.39 & 20.21 & 15.56 & 52.28 & 38.06 & 32.04 & 54.75 & 41.39 & 35.11 \\
& Sep-Attack & 89.88$^\ast$ & 82.60$^\ast$ & 78.82$^\ast$ & 42.92 & 27.04 & 20.65 & 54.46 & 40.12 & 33.46 & 55.88 & 41.30 & 35.18 \\
            &  Co-Attack  & 87.83$^\ast$ & 80.16$^\ast$ & 75.98$^\ast$ & 43.09 & 27.32 & 21.35 & 54.75 & 40.00 & 33.81 & 55.64 & 41.48 & 35.28 \\
			& \cellcolor{gray! 20}SGA  & \cellcolor{gray! 20}\textbf{96.95$\pm$0.08$^\ast$} & \cellcolor{gray! 20}\textbf{93.44$\pm$0.04$^\ast$} & \cellcolor{gray! 20}\textbf{91.00$\pm$0.06$^\ast$} & \cellcolor{gray! 20}\textbf{65.38$\pm$0.08} & \cellcolor{gray! 20}\textbf{47.61$\pm$0.07} & \cellcolor{gray! 20}\textbf{38.96$\pm$0.07} & \cellcolor{gray! 20}\textbf{65.25$\pm$0.09} & \cellcolor{gray! 20}\textbf{50.42$\pm$0.08} & \cellcolor{gray! 20}\textbf{43.47$\pm$0.12} & \cellcolor{gray! 20}\textbf{66.52$\pm$0.18} & \cellcolor{gray! 20}\textbf{52.44$\pm$0.28} & \cellcolor{gray! 20}\textbf{45.05$\pm$0.07} \\
			\midrule
			\multirow{5}{*}{\rotatebox[origin=c]{0}{\textbf{TCL}}} 
            & PGD & 16.52 & 8.40 & 5.61 & 69.53$^\ast$ & 60.88$^\ast$ & 57.56$^\ast$ & 22.28 & 12.20 & 9.10 & 23.12 & 12.77 & 9.49 \\
            & BERT-Attack & 45.92 & 30.40 & 23.89 & 48.48$^\ast$ & 31.48$^\ast$ & 24.47$^\ast$ & 58.80 & 43.10 & 36.68 & 61.26 & 46.14 & 39.54 \\
            & Sep-Attack & 52.97 & 36.33 & 28.97 & 78.97$^\ast$ & 69.79$^\ast$ & 65.71$^\ast$ & 60.13 & 44.13 & 37.32 & 61.26 & 45.99 & 38.97 \\
            &  Co-Attack  & 57.09 & 39.85 & 32.00 & 91.39$^\ast$ & 83.16$^\ast$ & 78.05$^\ast$ & 60.46 & 45.16 & 37.73 & 62.49 & 46.61 & 39.74 \\
			& \cellcolor{gray! 20}SGA  & \cellcolor{gray! 20}\textbf{73.30$\pm$0.04} & \cellcolor{gray! 20}\textbf{58.40$\pm$0.09} & \cellcolor{gray! 20}\textbf{50.96$\pm$0.17} & \cellcolor{gray! 20}\textbf{99.15$\pm$0.03$^\ast$} & \cellcolor{gray! 20}\textbf{98.17$\pm$0.02$^\ast$} & \cellcolor{gray! 20}\textbf{97.34$\pm$0.01$^\ast$} & \cellcolor{gray! 20}\textbf{63.99$\pm$0.16} & \cellcolor{gray! 20}\textbf{49.87$\pm$0.09} & \cellcolor{gray! 20}\textbf{42.46$\pm$0.10} & \cellcolor{gray! 20}\textbf{65.70$\pm$0.19} & \cellcolor{gray! 20}\textbf{51.45$\pm$0.06} & \cellcolor{gray! 20}\textbf{44.64$\pm$0.06}  \\
			\midrule
			\multirow{5}{*}{\rotatebox[origin=c]{0}{\textbf{CLIP$_{\rm ViT}$}}} 
            & PGD & 10.75 & 4.64 & 2.91 & 13.74 & 6.77 & 4.32 & 66.85$^\ast$ & 51.80$^\ast$ & 46.02$^\ast$ & 11.34 & 6.50 & 4.66 \\
& BERT-Attack & 29.74 & 18.13 & 13.73 & 29.61 & 16.91 & 12.66 & 51.68$^\ast$ & 37.12$^\ast$ & 31.02$^\ast$ & 53.72 & 40.13 & 34.32 \\
& Sep-Attack & 34.61 & 21.00 & 16.15 & 36.84 & 22.63 & 17.03 & 77.94$^\ast$ & 66.77$^\ast$ & 60.69$^\ast$ & 49.76 & 37.51 & 31.74 \\
           &  Co-Attack  & 42.67 & 27.20 & 21.46 & 44.69 & 29.42 & 22.85 & 98.80$^\ast$ & 96.83$^\ast$ & 95.33$^\ast$ & 62.51 & 49.48 & 42.63   \\
			& \cellcolor{gray! 20}SGA  & \cellcolor{gray! 20}\textbf{44.64$\pm$0.00} & \cellcolor{gray! 20}\textbf{28.66$\pm$0.13} & \cellcolor{gray! 20}\textbf{22.64$\pm$0.09} & \cellcolor{gray! 20}\textbf{47.76$\pm$0.25} & \cellcolor{gray! 20}\textbf{32.30$\pm$0.04} & \cellcolor{gray! 20}\textbf{25.70$\pm$0.04} & \cellcolor{gray! 20}\textbf{99.79$\pm$0.00$^\ast$} & \cellcolor{gray! 20}\textbf{99.37$\pm$0.01$^\ast$} & \cellcolor{gray! 20}\textbf{98.94$\pm$0.07$^\ast$} & \cellcolor{gray! 20}\textbf{65.83$\pm$0.35} & \cellcolor{gray! 20}\textbf{53.58$\pm$0.25} & \cellcolor{gray! 20}\textbf{46.84$\pm$0.16}  \\
			\midrule
			\multirow{5}{*}{\rotatebox[origin=c]{0}{\textbf{CLIP$_{\rm CNN}$}}} 
            & PGD & 10.62 & 4.51 & 2.76 & 13.65 & 6.39 & 4.32 & 10.70 & 6.20 & 4.52 & 84.20$^\ast$ & 73.64$^\ast$ & 67.86$^\ast$ \\
& BERT-Attack & 34.64 & 21.13 & 16.25 & 29.61 & 16.91 & 12.66 & 57.49 & 42.73 & 36.23 & 62.17$^\ast$ & 47.80$^\ast$ & 40.79$^\ast$ \\
& Sep-Attack & 39.29 & 24.04 & 18.83 & 41.51 & 26.13 & 20.17 & 57.11 & 41.89 & 35.55 & 92.49$^\ast$ & 85.84$^\ast$ & 81.66$^\ast$ \\
            &  Co-Attack  & 41.97 & 26.62 & 20.91 & 43.72 & 28.62 & 22.35 & 58.90 & 45.22 & 38.72 & 98.56$^\ast$ & 96.86$^\ast$ & 95.55$^\ast$ \\
			& \cellcolor{gray! 20}SGA  & \cellcolor{gray! 20}\textbf{43.00$\pm$0.01} & \cellcolor{gray! 20}\textbf{27.64$\pm$0.04} & \cellcolor{gray! 20}\textbf{21.74$\pm$0.00} & \cellcolor{gray! 20}\textbf{45.95$\pm$0.23} & \cellcolor{gray! 20}\textbf{30.57$\pm$0.00} & \cellcolor{gray! 20}\textbf{24.27$\pm$0.22} & \cellcolor{gray! 20}\textbf{60.77$\pm$0.02} & \cellcolor{gray! 20}\textbf{46.99$\pm$0.11} & \cellcolor{gray! 20}\textbf{40.49$\pm$0.16} & \cellcolor{gray! 20}\textbf{99.80$\pm$0.03$^\ast$} & \cellcolor{gray! 20}\textbf{99.29$\pm$0.06$^\ast$} & \cellcolor{gray! 20}\textbf{98.77$\pm$0.06$^\ast$} \\
			\bottomrule[0.3mm]
	\end{tabular}}
	\caption{\textbf{Attack success rate ($\%$)} of four VLP models under existing adversarial attacks and SGA.
	The source column indicates the source models used to generate the adversarial data on MSCOCO.
	 $^\ast$ indicates white-box attacks. A higher ASR indicates better adversarial transferability.}
	\vspace{-8pt}
	\label{tab:supp_t13_SGA_coco}
\end{sidewaystable*}

\subsection{Transferability Analysis}
\label{sec:supp_b_ana}
Table \ref{tab:supp_t8_exp_sep_attack_albef}, Table \ref{tab:supp_t9_exp_sep_attack_tcl}, Table \ref{tab:supp_t10_exp_sep_attack_clip}, and Table \ref{tab:supp_t11_exp_co_attack} show adversarial transferability among different VLP models and configurations under Sep-Attack and Co-Attack.
We report the attack success rates of the adversarial examples generated by the source model to attack the target models. 

Some observations on adversarial transferability are summarized below:
\begin{itemize}
  \item For all VLP models, attacking two modalities simultaneously shows better adversarial transferability than only attacking a single modality. This is consistent with the observation in \cite{Zhang2022Co-attack} for the white-box setting.
  \item Even though models with exact same architectures but with different pretrain objectives (\textit{e.g.}, ALBEF and TCL), the adversarial examples cannot directly pass through another model with a similar success attack rate. 
  \item The adversarial transferability from fused VLP models to aligned VLP models is higher than that from backward (\textit{e.g.}, from ALBEF or TCL to CLIP-ViT and CLIP-CNN). 
  \item Although ALBEF, TCL, and CLIP-ViT are using ViT as image-encoders, the adversarial transferability from ALBEF or TCL to CLIP-CNN will be higher than that of CLIP-ViT; similarly, the adversarial transferability of CLIP-ViT to CLIP-CNN is higher than that of CLIP-CNN to CLIP-ViT. 
\end{itemize}

\begin{table*}[t]
\begin{center}
\small
\renewcommand\arraystretch{0.81}
  \setlength{\tabcolsep}{2.5mm}{
\begin{tabular}{l|l|l|l|ccc|ccc}
        \toprule[0.3mm]
        \multirow{2}{*}{\textbf{\fontsize{10pt}{\baselineskip}\selectfont{Source}}} &
        \multirow{2}{*}{\textbf{\fontsize{10pt}{\baselineskip}\selectfont{Attack}}} &
        \multirow{2}{*}{\textbf{\fontsize{10pt}{\baselineskip}\selectfont{Target}}} & \multirow{2}{*}{\textbf{\fontsize{10pt}{\baselineskip}\selectfont{Method}}} & \multicolumn{3}{c|}{\textbf{\fontsize{10pt}{\baselineskip}\selectfont{Image-to-Text}}} & \multicolumn{3}{c}{\textbf{\fontsize{10pt}{\baselineskip}\selectfont{Text-to-Image}}}   \\
        &  &  &   & R@1  & R@5  & R@10    & R@1  & R@5  & R@10 \\
      \hline
      \multirow{16}{*}{ALBEF} & \multirow{8}{*}{\texttt{\textbf{Text}@Multi}}  & \multirow{2}{*}{ALBEF} 
            & Co-Attack   & 9.18$^\ast$ & 1.50$^\ast$ & 1.00$^\ast$ & 21.70$^\ast$ & 11.96$^\ast$ & 9.22$^\ast$  \\
      &  &  & SGA  & \textbf{13.03$^\ast$} & \textbf{2.71$^\ast$} & \textbf{1.40$^\ast$} & \textbf{26.17$^\ast$} & \textbf{14.17$^\ast$} & \textbf{10.76$^\ast$}  \\
      \cline{3-10}
      &  & \multirow{2}{*}{TCL}  & Co-Attack  & 9.38 & 1.31 & 0.30 & 20.40 & 9.74 & 6.80 \\
      &  &                       & SGA & \textbf{12.64} & \textbf{2.01} & \textbf{0.80} & \textbf{26.43} & \textbf{13.69} & \textbf{9.32}\\
      \cline{3-10}
      &  & \multirow{2}{*}{CLIP$_{\rm ViT}$} & Co-Attack  & 20.98 & 7.79 & 4.57 & 31.73 & 19.13 & 14.63  \\
      &  &                                   & SGA & \textbf{27.24} & \textbf{11.32} & \textbf{7.52} & \textbf{36.82} & \textbf{22.10} & \textbf{16.99}   \\ 
      \cline{3-10}
      &  & \multirow{2}{*}{CLIP$_{\rm CNN}$} & Co-Attack  & 22.48 & 7.40 & 4.12 & 31.94 & 21.36 & 15.59  \\
      &  &                                   & SGA & \textbf{27.97} & \textbf{13.85} & \textbf{7.62} & \textbf{37.77} & \textbf{24.82} & \textbf{18.46}   \\        
      \cline{2-10}
      
      &  \multirow{8}{*}{\texttt{\textbf{Image}@Multi}}       & \multirow{2}{*}{ALBEF} 
                & Co-Attack  & 75.50$^\ast$ & 59.22$^\ast$ & 53.30$^\ast$ & 83.63$^\ast$ & 75.14$^\ast$ & 70.32$^\ast$  \\
      &  &      & SGA & \textbf{90.82$^\ast$} & \textbf{83.27$^\ast$} & \textbf{79.00$^\ast$} & \textbf{90.08$^\ast$} & \textbf{83.35$^\ast$} & \textbf{79.32$^\ast$}    \\
      \cline{3-10}
      &  & \multirow{2}{*}{TCL}  & Co-Attack  & 4.64 & 1.21 & 0.50 & 11.33 & 3.72 & 2.25 \\
      &  &                             & SGA & \textbf{21.18} & \textbf{9.15} & \textbf{5.91} & \textbf{28.00} & \textbf{13.50} & \textbf{9.16}  \\
      \cline{3-10}
      &  & \multirow{2}{*}{CLIP$_{\rm ViT}$} & Co-Attack  & 7.24 & 1.97 & 0.51 & 13.53 & 5.23 & 3.01   \\
      &  &                                         & SGA & \textbf{10.92} & \textbf{3.53} & \textbf{1.52} & \textbf{16.72} & \textbf{6.70} & \textbf{4.34}   \\   
      \cline{3-10}
      &  & \multirow{2}{*}{CLIP$_{\rm CNN}$} & Co-Attack  & 10.09 & 2.85 & 1.65 & 15.27 & 6.11 & 3.52   \\
      &  &                                         & SGA & \textbf{12.52} & \textbf{3.91} & \textbf{2.47} & \textbf{17.77} & \textbf{7.44} & \textbf{4.65}   \\    
        \hline
        
      \multirow{16}{*}{TCL} & \multirow{8}{*}{\texttt{\textbf{Text}@Multi}}  & \multirow{2}{*}{ALBEF} 
            & Co-Attack & 13.24 & 2.61 & 1.20 & 27.13 & 15.16 & 11.28 \\
      &  &  & SGA & \textbf{10.84} & \textbf{2.71} & \textbf{0.90} & \textbf{24.77} & \textbf{12.22} & \textbf{9.30} \\
      \cline{3-10}
      &  & \multirow{2}{*}{TCL}  & Co-Attack  & 12.86 & 2.81 & 1.00 & 30.33 & 15.32 & 10.89  \\
      &  &                             & SGA & \textbf{13.38$^\ast$} & \textbf{3.72$^\ast$} & \textbf{1.00$^\ast$} & \textbf{27.17$^\ast$} & \textbf{14.06$^\ast$} & \textbf{10.07$^\ast$}  \\
      \cline{3-10}
      &  & \multirow{2}{*}{CLIP$_{\rm ViT}$} & Co-Attack  & 25.28 & 9.87 & 5.79 & 37.11 & 22.85 & 17.25 \\
      &  &                                         & SGA & \textbf{27.98} & \textbf{12.05} & \textbf{7.32} & \textbf{37.69} & \textbf{22.73} & \textbf{17.31}  \\ 
      \cline{3-10}
      &  & \multirow{2}{*}{CLIP$_{\rm CNN}$} & Co-Attack  & 26.18 & 11.21 & 5.25 & 37.84 & 24.65 & 18.71   \\
      &  &                                         & SGA  & \textbf{30.40} & \textbf{13.85} & \textbf{8.14} & \textbf{37.77} & \textbf{25.14} & \textbf{19.41}    \\        
      \cline{2-10}
      
      &  \multirow{8}{*}{\texttt{\textbf{Image}@Multi}}       & \multirow{2}{*}{ALBEF} 
               & Co-Attack  & 5.94 & 1.60 & 0.80 & 12.16 & 3.79 & 2.26  \\
      &  &     & SGA & \textbf{27.11} & \textbf{13.93} & \textbf{9.80} & \textbf{34.49} & \textbf{19.24} & \textbf{13.67}   \\
      \cline{3-10}
      &  & \multirow{2}{*}{TCL}  & Co-Attack  & 72.50 & 55.98 & 46.49 & 79.26 & 64.65 & 56.99   \\
      &  &                       & SGA & \textbf{96.00$^\ast$} & \textbf{92.16$^\ast$} & \textbf{89.28$^\ast$} & \textbf{96.86$^\ast$} & \textbf{92.70$^\ast$} & \textbf{90.19$^\ast$}  \\
      \cline{3-10}
      &  & \multirow{2}{*}{CLIP$_{\rm ViT}$} & Co-Attack  & 7.85 & 1.97 & 0.61 & 13.43 & 5.37 & 3.31   \\
      &  &                                         & SGA & \textbf{10.92} & \textbf{3.53} & \textbf{1.52} & \textbf{16.88} & \textbf{7.15} & \textbf{4.62}    \\   
      \cline{3-10}
      &  & \multirow{2}{*}{CLIP$_{\rm CNN}$} & Co-Attack  & 9.71 & 2.85 & 1.65 & 15.44 & 5.70 & 3.37  \\
      &  &                                         & SGA & \textbf{13.15} & \textbf{4.97} & \textbf{2.37} & \textbf{18.56} & \textbf{7.56} & \textbf{5.13}     \\    
        \hline

      \multirow{16}{*}{CLIP$_{\rm ViT}$} & \multirow{8}{*}{\texttt{\textbf{Text}@Multi}}  & \multirow{2}{*}{ALBEF} 
            & Co-Attack  & 7.61 & 1.00 & 0.30 & 19.97 & 9.58 & 6.59 \\
      &  &  & SGA & \textbf{8.13} & \textbf{1.20} & \textbf{0.40} & \textbf{19.50} & \textbf{8.76} & \textbf{6.59}  \\
      \cline{3-10}
      &  & \multirow{2}{*}{TCL}  & Co-Attack  & 8.43 & 0.90 & 0.30 & 20.90 & 9.96 & 7.03   \\
      &  &                             & SGA & \textbf{8.96} & \textbf{1.01} & \textbf{0.30} & \textbf{21.64} & \textbf{10.59} & \textbf{7.88}    \\
      \cline{3-10}
      &  & \multirow{2}{*}{CLIP$_{\rm ViT}$} & Co-Attack   & 28.34 & 11.73 & 6.81 & 38.89 & 24.08 & 17.42  \\
      &  &                                         & SGA & \textbf{31.78$^\ast$} & \textbf{15.16$^\ast$} & \textbf{8.43$^\ast$} & \textbf{39.43$^\ast$} & \textbf{25.58$^\ast$} & \textbf{19.30$^\ast$}  \\ 
      \cline{3-10}
      &  & \multirow{2}{*}{CLIP$_{\rm CNN}$} & Co-Attack  & 29.89 & 11.52 & 5.87 & 37.36 & 24.97 & 18.62 \\
      &  &                                         & SGA & \textbf{29.89} & \textbf{12.37} & \textbf{6.90} & \textbf{36.40} & \textbf{23.13} & \textbf{18.48}    \\        
      \cline{2-10}
      
      &  \multirow{8}{*}{\texttt{\textbf{Image}@Multi}}       & \multirow{2}{*}{ALBEF} 
              & Co-Attack  & 2.50 & 0.60 & 0.20 & 5.80 & 1.78 & 1.11 \\
      &  &     & SGA & \textbf{3.86} & \textbf{0.70} & \textbf{0.30} & \textbf{7.69} & \textbf{2.73} & \textbf{1.52}  \\
      \cline{3-10}
      &  & \multirow{2}{*}{TCL}  & Co-Attack  & 5.27 & 0.40 & 0.20 & 9.12 & 2.75 & 1.48  \\
      &  &                             & SGA & \textbf{6.43} & \textbf{0.60} & \textbf{0.20} & \textbf{10.93} & \textbf{3.47} & \textbf{2.05}   \\
      \cline{3-10}
      &  & \multirow{2}{*}{CLIP$_{\rm ViT}$} & Co-Attack  & 87.73 & 78.09 & 72.05 & 91.72 & 83.32 & 78.67 \\
      &  &                                         & SGA & \textbf{94.11$^\ast$} & \textbf{88.89$^\ast$} & \textbf{83.64$^\ast$} & \textbf{95.91$^\ast$} & \textbf{90.10$^\ast$} & \textbf{85.98$^\ast$}      \\   
      \cline{3-10}
      &  & \multirow{2}{*}{CLIP$_{\rm CNN}$} & Co-Attack  & 7.66 & 1.90 & 1.44 & 9.37 & 3.90 & 2.53   \\
      &  &                                         & SGA & \textbf{11.24} & \textbf{5.39} & \textbf{2.68} & \textbf{15.68} & \textbf{6.88} & \textbf{5.08}   \\    
        \hline

      \multirow{16}{*}{CLIP$_{\rm CNN}$} & \multirow{8}{*}{\texttt{\textbf{Text}@Multi}}  & \multirow{2}{*}{ALBEF} 
           & Co-Attack  & 7.72 & 0.90 & 0.50 & 20.79 & 9.84 & 6.98 \\
      &  &  & SGA & \textbf{7.82} & \textbf{1.30} & \textbf{0.60} & \textbf{19.93} & \textbf{9.74} & \textbf{7.16} \\
      \cline{3-10}
      &  & \multirow{2}{*}{TCL}  & Co-Attack  & 9.69 & 1.31 & 0.30 & 21.67 & 10.73 & 7.49   \\
      &  &                             & SGA & \textbf{9.59} & \textbf{1.91} & \textbf{0.60} & \textbf{21.88} & \textbf{10.96} & \textbf{7.74}     \\
      \cline{3-10}
      &  & \multirow{2}{*}{CLIP$_{\rm ViT}$} & Co-Attack  & 26.99 & 11.11 & 6.81 & 37.37 & 23.48 & 17.64   \\
      &  &                                         & SGA & \textbf{26.50} & \textbf{11.63} & \textbf{6.40} & \textbf{37.66} & \textbf{22.89} & \textbf{17.01}     \\ 
      \cline{3-10}
      &  & \multirow{2}{*}{CLIP$_{\rm CNN}$} & Co-Attack  & 30.40 & 13.11 & 7.21 & 40.03 & 26.79 & 20.74   \\
      &  &                                         & SGA & \textbf{36.27$^\ast$} & \textbf{17.34$^\ast$} & \textbf{11.02$^\ast$} & \textbf{44.29$^\ast$} & \textbf{29.16$^\ast$} & \textbf{22.82$^\ast$}      \\        
      \cline{2-10}
      
      &  \multirow{8}{*}{\texttt{\textbf{Image}@Multi}}       & \multirow{2}{*}{ALBEF} 
               & Co-Attack  & 1.98 & 0.30 & 0.20 & 5.12 & 1.42 & 0.91  \\
      &  &     & SGA  & \textbf{2.09} & \textbf{0.60} & \textbf{0.20} & \textbf{6.20} & \textbf{1.70} & \textbf{1.19}    \\
      \cline{3-10}
      &  & \multirow{2}{*}{TCL}  & Co-Attack  & 4.74 & 0.50 & 0.10 & 7.95 & 2.32 & 1.42 \\
      &  &                             & SGA & \textbf{4.85} & \textbf{0.70} & \textbf{0.30} & \textbf{9.19} & \textbf{2.63} & \textbf{1.73}      \\
      \cline{3-10}
      &  & \multirow{2}{*}{CLIP$_{\rm ViT}$} & Co-Attack  & 1.84 & 0.10 & 0.30 & 5.51 & 2.50 & 1.02    \\
      &  &                                         & SGA & \textbf{3.19} & \textbf{1.77} & \textbf{0.81} & \textbf{9.34} & \textbf{4.56} & \textbf{2.35}     \\   
      \cline{3-10}
      &  & \multirow{2}{*}{CLIP$_{\rm CNN}$} & Co-Attack & 88.12 & 79.70 & 74.87 & 93.69 & 87.66 & 83.03 \\
      &  &                                         & SGA & \textbf{92.46$^\ast$} & \textbf{86.68$^\ast$} & \textbf{81.98$^\ast$} & \textbf{96.64$^\ast$} & \textbf{91.78$^\ast$} & \textbf{87.87$^\ast$}  \\    
        \bottomrule[0.3mm]
    \end{tabular}}
\end{center}
\vspace{-10pt}
\caption{
\textbf{Attack success rates ($\%$)} on four VLP models under Co-Attack \cite{Zhang2022Co-attack} and SGA with different single adversarial input modalities. The adversaries are crafted on Flickr30K. $^\ast$ indicates white-box attacks.}
\label{tab:supp_t14_exp_unimodal}
\end{table*}

\subsection{Main Results}
\label{sec:supp_b_main_res}
We present a thorough analysis of the performance of our proposed high transferable multimodal attack method, SGA, on the popular benchmark datasets Flickr30K and MSCOCO. 
The experimental results are summarized in Table \ref{tab:supp_t12_SGA_flickr} and Table \ref{tab:supp_t13_SGA_coco}, providing a clear comparison between the performance of our SGA and existing multimodal attack methods across different attack scenarios. 
As we can see, our proposed SGA outperforms the state-of-the-art in all white-box and black-box settings.
Moreover, as illustrated in Table \ref{tab:supp_t14_exp_unimodal}, we conduct extensive experiments on Flickr30K under a unimodal scenario, with perturbed input in either the image or text modality. 
Empirical evidence suggests that even in scenarios where only query data are accessible, the performance of SGA consistently surpasses that of existing methods.

Our results suggest that the proposed SGA can serve as a promising method for evaluating the robustness of multimodal models and improving their security in real-world applications.

\subsection{Ablation Study}
\label{sec:supp_b_ablation}

This section presents the ablation experiments on the augmented multimodal data and the iterative strategy sequence of SGA. To provide a thorough analysis, detailed experimental results are presented and discussed.

\paragraph{Iterative Strategy.}
In this study, we generate adversarial examples through cross-modal guidance. 
This allows for the disruption of multimodal interactions through the collaborative generation of perturbations. 
Notably, our process follows a ``text-image-text" (t-i-t) pipeline.

We have conducted additional experiments to evaluate the effectiveness of our attack strategy. 
As shown in Table \ref{tab:supp_t15_ablation_reverse_multi_alter}, an interesting observation is that reversing the ``t-i-t" pipeline does not significantly impact the results. 
Furthermore, although adding one iteration (t-i-t-i-t) slightly enhances performance, it also doubles the computational cost. 
This suggests that our SGA is not sensitive to the exact order of the pipeline, but rather benefits from cross-modal guidance.

\paragraph{Multi-scale Image Set.}
In SGA, an augmented image set is used to generate adversarial data based on the scale-invariant property of deep learning models.
To verify the effectiveness of the augmented image set, we choose different scale ranges to build the image sets and evaluate the adversarial transferability.
As presented in Table \ref{tab:supp_t16_ablation_img}, there exists a positive correlation between transferability and the scale range, with the highest transferability observed at a scale range of $[0.50,1.50]$ with a step size of 0.25.
The experimental results show that the augmented image set plays a crucial role in increasing the transferability of the generated adversarial data.

\begin{table*}[t]
\begin{center}
\renewcommand\arraystretch{1.1}
\setlength{\tabcolsep}{3mm}
		\scalebox{0.97}[0.97]{
		    \begin{tabular}{c|ccc|ccc}
        \toprule
        \multirow{2}{*}{\textbf{\fontsize{10pt}{\baselineskip}\selectfont{Iterative Strategy}}} & \multicolumn{3}{c|}{\textbf{\fontsize{10pt}{\baselineskip}\selectfont{Image-to-Text}}} & \multicolumn{3}{c}{\textbf{\fontsize{10pt}{\baselineskip}\selectfont{Text-to-Image}}}   \\
            & R@1  & R@5  & R@10    & R@1  & R@5  & R@10 \\
      \midrule
    t-i-t       & 45.42     & 24.93    & 16.48     & 55.25     & 36.01     & 27.25     \\
    i-t-i       & 45.84     & 26.43    & 18.24     & 56.45     & 36.39     & 27.60     \\
    t-i-t-i-t   & 48.37     & 26.63    & 19.44     & 57.19     & 38.08     & 28.72  \\
        \bottomrule
    \end{tabular}}
\end{center}
\caption{ \textbf{Ablation experiment on different iterative strategies.} The dataset is Flickr30K. The source model is ALBEF and the target model is TCL. Attack success rates ($\%$) are utilized to measure the adversarial transferability. }
\vspace{-5pt}
\label{tab:supp_t15_ablation_reverse_multi_alter}
\end{table*}
\begin{table*}[t]
\begin{center}
\renewcommand\arraystretch{1.1}
\setlength{\tabcolsep}{3mm}
		\scalebox{0.97}[0.97]{
		    \begin{tabular}{c|ccc|ccc}
        \toprule
        \multirow{2}{*}{\textbf{\fontsize{10pt}{\baselineskip}\selectfont{Scales}}} & \multicolumn{3}{c|}{\textbf{\fontsize{10pt}{\baselineskip}\selectfont{Image-to-Text}}} & \multicolumn{3}{c}{\textbf{\fontsize{10pt}{\baselineskip}\selectfont{Text-to-Image}}}   \\
            & R@1  & R@5  & R@10    & R@1  & R@5  & R@10 \\
      \midrule
      $[1.00]$                & 34.04    & 13.17      & 8.62      & 44.12     & 25.95     & 19.25     \\
      $[0.75, 1.00, 1.25]$                 & 44.57    & 22.70      & 14.63     & 54.55     & 34.36     & 26.22  \\
      $[0.50, 0.75, 1.00, 1.25,1.50]$      & 45.94    & 24.82      & 16.13     & 55.21     & 35.99     & 27.15   \\
      $[0.25, 0.50, 0.75, 1.00, 1.25,1.50, 1.75]$   & 44.15   & 24.22  & 16.13  & 55.10    & 35.35  & 26.81  \\
        \bottomrule
    \end{tabular}}
\end{center}
\caption{ \textbf{Ablation experiment on the image set.} The dataset is Flickr30K. The source model is ALBEF and the target model is TCL. Attack success rates ($\%$) are utilized to measure the adversarial transferability. }
\vspace{-5pt}
\label{tab:supp_t16_ablation_img}
\end{table*}
\begin{table*}[t]
\begin{center}
\renewcommand\arraystretch{1.1}
\setlength{\tabcolsep}{3mm}
		\scalebox{0.97}[0.97]{
		    \begin{tabular}{c|ccc|ccc}
        \toprule
        \multirow{2}{*}{\textbf{\fontsize{10pt}{\baselineskip}\selectfont{Number of Captions}}} & \multicolumn{3}{c|}{\textbf{\fontsize{10pt}{\baselineskip}\selectfont{Image-to-Text}}} & \multicolumn{3}{c}{\textbf{\fontsize{10pt}{\baselineskip}\selectfont{Text-to-Image}}}   \\
            & R@1  & R@5  & R@10    & R@1  & R@5  & R@10 \\
      \midrule
      1                 & 40.04     & 18.99         & 12.53     & 51.14     & 30.93     & 23.17     \\
      2                 & 45.52     & 22.51         & 15.13     & 54.69     & 33.45     & 25.28  \\
      3                 & 45.84     & 23.82         & 15.43     & 54.67     & 34.69     & 26.58   \\
      4                 & 46.05     & 25.03         & 16.23     & 55.16     & 35.66     & 27.13         \\
      5                 & 45.94     & 24.82         & 16.13     & 55.21     & 35.99     & 27.15         \\
        \bottomrule
    \end{tabular}}
\end{center}
\vspace{-8pt}
\caption{ \textbf{Ablation experiment on the caption set. }The dataset is Flickr30K. The source model is ALBEF and the target model is TCL. Attack success rates ($\%$) are utilized to measure the adversarial transferability. }
\vspace{-5pt}
\label{tab:supp_t17_ablation_txt}
\end{table*}

\paragraph{Multi-pair Caption Set.}
The proposed SGA involves augmenting the original caption into a caption set for the purpose of generating adversarial data. 
To determine the effectiveness of the augmented caption set, various numbers of captions are utilized to construct the caption sets, and the transferability of the resulting adversarial data is evaluated. 
As illustrated in Table \ref{tab:supp_t17_ablation_txt}, the use of multiple captions in the process of crafting adversarial data is observed to have a significant positive impact on adversarial transferability. 
Experimental results demonstrate that the augmented caption set also helps enhance the transferability of the generated adversarial data.

\subsection{Visualization}
\label{sec:supp_b_vis}
\begin{figure*}[t]
\begin{center}
   \includegraphics[width=0.9\linewidth]{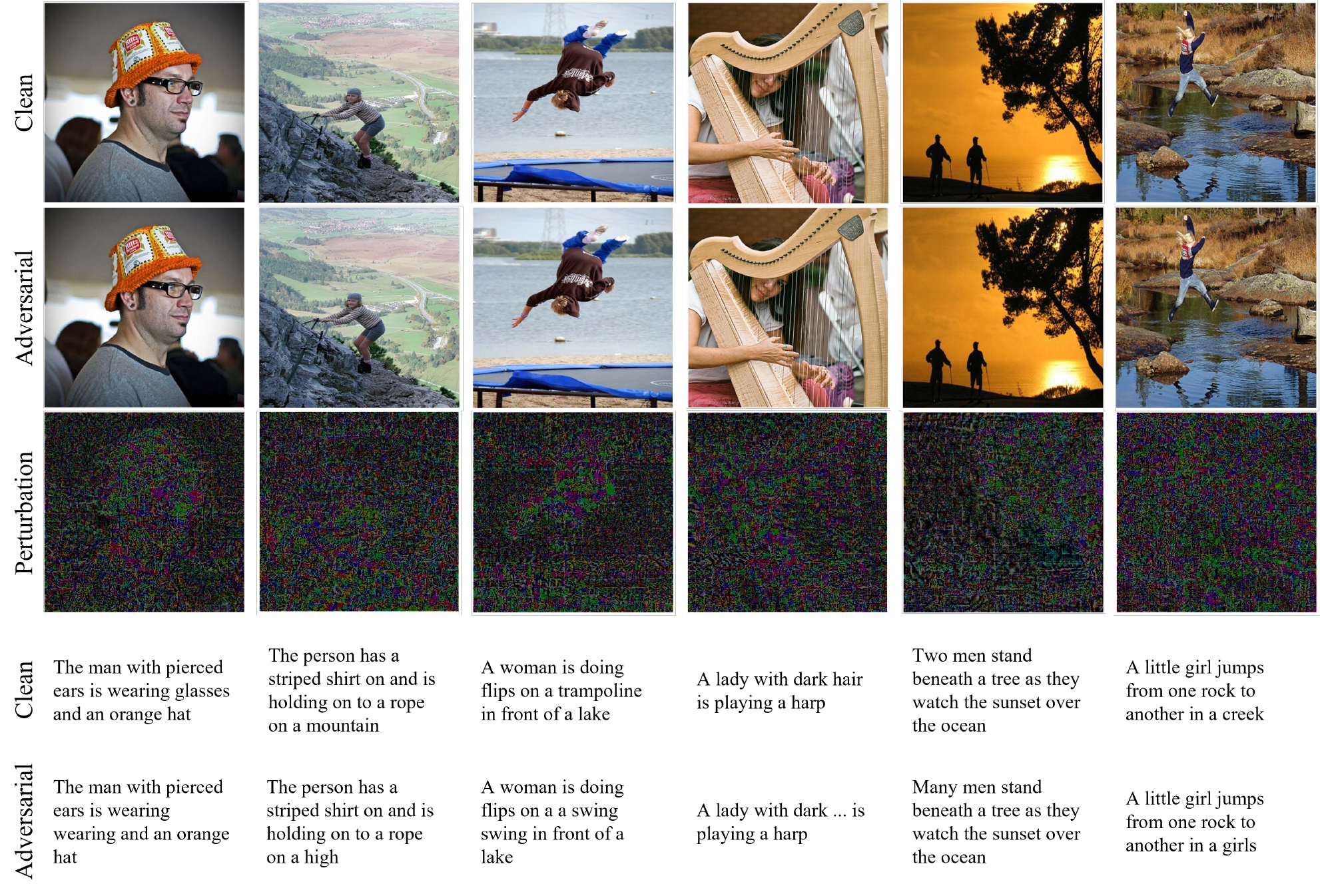}
\end{center}
\vspace{-10pt}
   \caption{\textbf{Visualization} of original images (\textbf{Upper}) and the corresponding adversarial examples (\textbf{Middle}) generated by our proposed SGA. Perturbations (\textbf{Lower}) are amplified by a factor of $50$ for better illustration.}
\label{fig:supp_f7_visualization}
\end{figure*}

Figure \ref{fig:supp_f7_visualization} depicts randomly selected original clean images and the corresponding adversarial examples, and such small perturbations are hard to be perceived.
We magnified the imperceptible perturbation by a factor of 50 for visualization.

\section{Algorithm}
\label{sec:supp_c}

\begin{algorithm}[t]
\caption{Set-level Guidance Attack}
\begin{algorithmic}
\State \textbf{Input:} Image encoder $f_I$, Text encoder $f_T$, Dataset $D$, Image-caption pair $(v,t)$, Image scale sets $S=\{s_1,s_2,...,s_N\}$, iteration steps $K$, number of paired captions $M$ 
\State \textbf{Output:} adversarial image $v'$, adversarial caption $t'$
\State Build caption set $\boldsymbol{t}=\{t_1,t_2,...,t_M\} \gets D$ \\
$/*$ Build adversarial caption set $\boldsymbol{t'}=\{t_1^{'},t_2^{'},...,t_M^{'}\}$ $*/$
    \For{$iter$ $i = 1, 2, ..., M$} 
        \State $t_{i}' = \mathop{\arg\max}\limits_{t_{i}'\in {B}[t_i,\epsilon_t]}-\frac{f_T(t_{i}')\cdot f_I(v)}{\Vert f_T(t_{i}') \Vert \Vert f_I(v) \Vert}$
    \EndFor \\
$/*$ Build image set $\boldsymbol{v}=\{v_1,v_2,...,v_N\}$ $*/$
\For{$iter$ $i = 1, 2, ..., N$} 
    \State $v_i = resize(v, s_i) + 0.05 \cdot \boldsymbol{N}(0,1)$
\EndFor \\
$/*$ Generate adversarial image $v'$ $*/$
\For{$iter$ $k = 1, 2, ..., K$}
    \State $v'=\mathop{\arg\max}\limits_{v'\in {B}[v,\epsilon_v]} -\sum_{i=1}^{M}\frac{f_T(t_{i}')}{\Vert f_T(t_{i}')\Vert}\sum_{v_i\in \boldsymbol{v}}\frac{f_I(v_i)}{\Vert f_I(v_i)\Vert}$
\EndFor \\
$/*$ Generate adversarial caption $t'$ $*/$
\State \quad \space $t' = \mathop{\arg\max}\limits_{t'\in {B}[t,\epsilon_t]}-\frac{f_T(t')\cdot f_I(v')}{\Vert f_T(t') \Vert \Vert f_I(v') \Vert}$
\end{algorithmic}
\label{alg:sga_algp}
\end{algorithm}

The detailed training process of our proposed SGA is described in Algorithm \ref{alg:sga_algp}.

\end{document}